\documentclass{article}

\usepackage{microtype}
\usepackage{graphicx}
\usepackage{subcaption}
\usepackage{booktabs} 

\usepackage{hyperref}

\usepackage[accepted]{icml2026}

\usepackage{amsmath}
\usepackage{amssymb}
\usepackage{mathtools}
\usepackage{amsthm}

\usepackage[capitalize,noabbrev]{cleveref}

\usepackage{arydshln}
\usepackage{xcolor}
\usepackage[table]{xcolor}
\definecolor{myblue}{rgb}{0, 0.5, 0.7}
\definecolor{mypurple}{rgb}{0.6, 0.2, 0.8}
\usepackage{multirow}
\usepackage{makecell}
\usepackage{pifont} 
\usepackage{bbding}       
\usepackage{fontawesome} 
\usepackage{longfbox}
\makeatletter
\newdimen\@tempdimd
\makeatother

\theoremstyle{plain}

\theoremstyle{definition}

\theoremstyle{remark}

\usepackage[textsize=tiny]{todonotes}

\icmltitlerunning{Weakly Supervised Cross-Modal Learning for 4D Radar Scene Flow Estimation}

\begin{document}

\twocolumn[
  \icmltitle{Weakly Supervised Cross-Modal Learning for 4D Radar Scene Flow Estimation}



  \icmlsetsymbol{equal}{*}

  \begin{icmlauthorlist}
   \icmlauthor{Jingyun Fu}{xxx}
   \icmlauthor{Zhiyu Xiang}{xxx,yyy}
   \icmlauthor{Na Zhao}{zzz}
  \end{icmlauthorlist}

  \icmlaffiliation{xxx}{Zhejiang University, Hangzhou, Zhejiang, China}
  \icmlaffiliation{yyy}{Zhejiang Provincial Key Laboratory of Multi-Modal Communication Networks and Intelligent Information Processing, Zhejiang, China}
  \icmlaffiliation{zzz}{Singapore University of Technology and Design, Singapore}

  \icmlcorrespondingauthor{Zhiyu Xiang}{xiangzy@zju.edu.cn}
  \icmlcorrespondingauthor{Na Zhao}{na\_zhao@sutd.edu.sg}

  \icmlkeywords{Scene Flow, 4D radar, Weak Supervision}

  \vskip 0.3in
]



\printAffiliationsAndNotice{}  

\begin{abstract}
  Due to the difficulty of obtaining ground-truth data for 4D radar scene flow estimation, previous methods typically rely on either self-supervised losses or cross-modal supervision using 3D LiDAR data, 2D images, and odometry. However, self-supervised approaches often yield suboptimal results due to radar's inherently low-fidelity measurements, while existing cross-modal supervised methods introduce complex multi-task architecture and require costly LiDAR sensors to generate pseudo radar scene flow labels from pretrained 3D tracking models. To overcome these limitations, we propose a task-specific iterative framework for weakly supervised radar scene flow learning, using only images and odometry for auxiliary supervision during training. Specially, we establish two novel instance-aware self-supervised losses by exploiting off-the-shelf 2D tracking and segmentation algorithms to obtain tracked instance masks, which are back-projected into 3D space to provide instance-level semantic guidance; for static regions, we integrate vehicle odometry with radar's intrinsic motion cues to construct a rigid static loss. Extensive experiments on the real-world View-of-Delft (VoD) dataset demonstrate that our method not only surpasses state-of-the-art cross-modal supervised approaches that rely on 3D multi-object tracking on dense LiDAR point clouds but also outperforms existing fully supervised scene flow estimation methods. The code is open-sourced at \href{https://github.com/FuJingyun/IterFlow}{https://github.com/FuJingyun/IterFlow}.
\end{abstract}

\section{Introduction}
\label{sec:intro}
Scene flow estimation aims to predict the 3D motion field between consecutive point cloud frames, providing fine-grained, class-agnostic motion cues for 3D dynamic environment understanding. It further supports downstream tasks in autonomous driving, such as motion segmentation, visual SLAM, detection and tracking. While LiDAR-based scene flow has become mainstream~\cite{liu2019flownet3d,puy2020flot, wei2021pv, kittenplon2021flowstep3d, zhang2024deflow,kim2025flow4d, khoche2025ssf,  lin2025voteflow}, the emerging 4D automotive radar is receiving increasing attention due to its lower cost, higher robustness under adverse weather conditions and the ability to measure extra radial relative velocity (RRV). However, since radar point clouds are significantly sparser and noisier than LiDAR, radar-based scene flow estimation is challenging and remains largely underexplored.

\begin{figure}[t!]
  \centering
\includegraphics[width=0.95\linewidth]{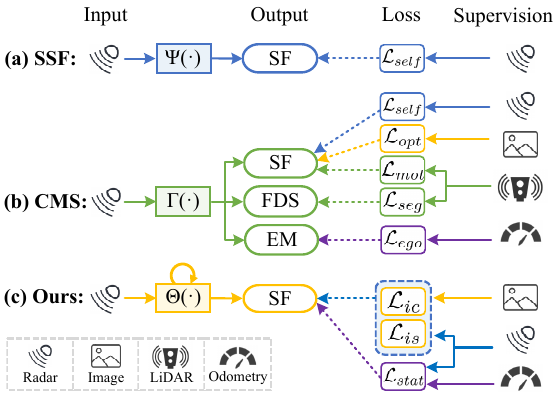}
   \caption{
   Comparison between existing \textit{self-supervised} (SSF) and \textit{cross-modal supervised} (CMS) radar scene flow estimation settings and our \textit{weakly supervised cross-modal} learning setting. \colorbox{gray!30}{SF}, \colorbox{gray!30}{FDS}, and \colorbox{gray!30}{EM} denote the predicted scene flow, foreground dynamic segmentation, and ego-motion, respectively. 
$\mathcal{L}_{self}$ is the self-supervised losses in \cite{ding2022self}; $\mathcal{L}_{opt}$, $\mathcal{L}_{mot}$, $\mathcal{L}_{seg}$ and $\mathcal{L}_{ego}$ are cross-modal losses proposed in ~\cite{ding2023hidden,zhai2025dmrflow,wu2025tars}, with supervision from 2D optical flow, 3D LiDAR-bassed pseudo scene flow label and FDS ground-truth, and odometry-based ego-motion. $\mathcal{L}_{ic}$ and $\mathcal{L}_{is}$ are our instance-aware losses, and $\mathcal{L}_{stat}$ is the rigid static loss.
   }
   \label{fig:method_comparison}
\end{figure}

Due to the high cost of obtaining pointwise ground-truth 3D flow annotations for radar scene flow estimation, recent studies have increasingly focused on \textit{label-free} 4D radar scene flow estimation. A straightforward attempt is to extend LiDAR-based self-supervised approaches, but the commonly used clustering strategies ~\cite{zhang2024seflow,lin2025voteflow} and Chamfer-guided~\cite{wu2020pointpwc,mittal2020just,pontes2020scene} pseudo correspondences are originally designed for high-performance LiDAR point clouds and cannot be easily applied to the challenging radar domain. To address the intractability of radar data, the pioneering RaFlow~\cite{ding2022self} introduces specialized self-supervised loss designs (Fig.~\ref{fig:method_comparison}a), using radar RRV as a supervision signal and relaxing LiDAR-oriented constraints to better adapt to radar conditions. However, the performance of RaFlow remains suboptimal due to the inherently low-fidelity nature of radar signals. Subsequent works~\cite{ding2023hidden, zhai2025dmrflow, wu2025tars} adopt multi-task network architectures to leverage cross-modal supervision signals (Fig.~\ref{fig:method_comparison}b), leading to notable gains in flow prediction accuracy. Nevertheless, these frameworks are highly complicated and require seven or more loss terms for network training, including all self-supervised losses from RaFlow~\cite{ding2022self} as well as additional cross-modal losses derived from 3D LiDAR, 2D optical flow, and vehicle odometry. Notably, although these methods avoid using ground-truth scene flow labels, they instead generate pseudo radar flow labels from 3D tracking results on dense LiDAR point clouds.

Given the high cost of high-performance LiDAR sensors, we propose a novel setting, weakly supervised cross-modal learning for 4D radar scene flow, that relies \textit{only on RGB images and odometry}, which are inexpensive sensors commonly available on consumer vehicles and robotic platforms. To the best of our knowledge, this is the \textit{first} work to leverage RGB images and odometry exclusively as auxiliary supervision for 4D radar scene flow during training (Fig.~\ref{fig:method_comparison}c), while requiring only radar point clouds as input at inference time. Under this new setting, we design a task-specific lightweight framework, termed \textbf{IterFlow}. Compared to previous CMS methods, IterFlow is computational efficient and effective, which is trained with only three complementary losses, including two novel self-supervised losses guided by 2D semantics extracted from images together with a rigid static loss based on odometry for weak supervision.

IterFlow excels by incorporating \textit{ball query-based cross-frame correlation} and an \textit{iterative scene flow refinement} scheme. When calculating the correlation between frames, the ball query-based grouping is more robust than traditional K-Nearest Neighbor (KNN) grouping~\cite{ward1963hierarchical,ester1996density} for radar point clouds, as it effectively avoids mismatches with distant points in sparse area. Moreover, iteratively updated scene flow enables IterFlow to 
refine coarse predicted flows from the initial steps and produce better scene flow estimation results than one-step inference approaches~\cite{cheng2022bi,ding2022self,ding2023hidden}.

As for loss design, we first leverage instance-level semantic guidance from 2D images to establish two novel instance-aware losses. Specifically, 2D tracking and segmentation results are back-projected from the image plane into 3D space, assigning instance labels to valid radar points. Based on these labels, we define an \textit{instance-aware Chamfer loss}, which computes the Chamfer distance only for point pairs belonging to the same instance across frames, and an \textit{instance-level spatial smoothness loss}, which enforces flow coherence within each object. By integrating 2D semantic correspondence with 3D geometric relationships between frames, 
these instance-aware losses effectively mitigate the mismatches and ambiguities caused by commonly used all-pair Chamfer calculations and KNN-based flow smoothing, ~\cite{wu2020pointpwc,ding2022self,ding2023hidden,zhai2025dmrflow,wu2025tars}
 yielding semantically enriched supervision for radar scene flow learning.
Additionally, we utilize vehicle odometry to construct a rigid static loss that supervises the flow estimation in static regions.

Our key contributions are summarized as follows:
\begin{itemize}
\item
To the best of our knowledge, we are the first to propose a task-specific cross-modal framework for 4D radar scene flow estimation, using only images and odometry for weak supervision during training. IterFlow is lightweight, featuring iterative flow refinement scheme and ball query-based cross-frame correlation, both tailored to the challenging radar domain.
\item
We design two novel instance-aware losses by leveraging semantic guidance from images, effectively mitigating the mismatch and ambiguity issues commonly encountered in existing self-supervised losses.
\item
Our proposed method achieves state-of-the-art performance in 4D radar scene flow estimation without requiring flow labels, surpassing previous cross-modal supervised radar-based methods that require pseudo 3D flow labels from LiDAR data and even outperforming fully supervised models.
\end{itemize}

\section{Related work}
\label{sec:related_work}

\noindent\textbf{Supervised LiDAR Scene Flow.}
With advances on deep neural network on point sets~\cite{qi2017pointnet,qi2017pointnet++,wu2019pointconv,zhao2021psˆ2,wang2024fly, sheng2025ct3d++, yuan2026graph}, 
LiDAR-based supervised scene flow learning has been extensively studied~\cite{liu2019flownet3d, puy2020flot, wei2021pv, cheng2022bi,fu2023pt, liu2024difflow3d, lin2025flowmamba} over the past few years. However, these methods focus on annotated synthetic datasets and confront the domain gap between real-world and synthetic data. To enable training on large-scale automotive datasets which do not provide ready-made scene flow labels, a series of follow-up works~\cite{jund2021scalable,zhang2024deflow, kim2025flow4d, khoche2025ssf, luo2025mambaflow, fu2026raliflow} utilize raw LiDAR point clouds and annotated tracking boxes to extract rigid flow labels for foreground objects. 
Using pillar-based~\cite{lang2019pointpillars} or voxel-based backbones~\cite{tang2020searching}, these methods can handle large-scale point clouds and have good real-time capability. Since LiDAR-based setups typically assume that ego motion is given, the performance of these methods is severely compromised when the vehicle's odometry is unknown. 
Additionally, although recent works~\cite{liu2022camliflow,peng2023delflow,wan2023rpeflow,zhou2024bring} have incorporated 2D images or event cameras with LiDAR point clouds for multi-modal scene flow estimation, they directly use data from other modalities as input to the network.

\noindent\textbf{Label-free LiDAR Scene Flow.}
To overcome the limitation of expensive 3D scene flow annotations, runtime optimization and self-supervised methods have witnessed remarkable progress in recent years. Although optimization-based methods~\cite{li2021neural,li2023fast,vedder2024neural} produce high quality flow predictions, they require excessive running time. In contrast, self-supervised deep learning methods enable faster inference and primarily rely on label-free proxy loss designs, including chamfer distance~\cite{pontes2020scene}, cycle-consistency\cite{mittal2020just}, local flow smoothness\cite{wu2020pointpwc}, and rigidity priors\cite{chodosh2024re, jiang20243dsflabelling, wang2025unsupervised}. Some advanced researches further apply point grouping~\cite{ward1963hierarchical,ester1996density} to cluster foreground points and construct self-supervised scene flow losses between point clusters~\cite{zhang2024seflow, lin2024icp, hoffmann2025floxels, lin2025voteflow}. However, LiDAR-based approaches naturally take advantage of rich spatial structural and geometric properties from high-resolution dense point clouds. Therefore, these methods cannot be easily applied to low-quality sparse radar point clouds.

\noindent\textbf{Radar Scene Flow.}
Although remarkable progress has been made in LiDAR-based studies, research on challenging radar scene flow estimation is still in the starting stage. 
Due to laborious 3D point cloud annotation, existing radar-based solutions prefer settings free of scene flow labels. RaFlow~\cite{ding2022self} first employs radar RRV as a direct supervision for flow prediction and also introduces relaxed chamfer loss and spatial flow smoothness loss to cope with difficult radar data. However, the sparse and noisy nature of radar point clouds hinders the reliability of such exclusive self-supervised learning. CMFlow~\cite{ding2023hidden} then proposes a multi-task framework to leverage additional cross-modal supervision from dense LiDAR point clouds, optical flows and odometry. Recently, radar-based network design is becoming increasingly complex and loss functions are constantly piling up to achieve higher performance. The latest ~\cite{zhai2025dmrflow} adds decoupled feature extraction branches and flow refinement modules into previous network; Wu et al.~\cite{wu2025tars} even integrates a pretrained 3D object detection network into the scene flow backbone. Unlike the mainstream trend, we strive to achieve effective radar scene flow estimation through lightweight task-specific network and concise loss design.

\section{Method}
\begin{figure*}[ht!]
  \centering
\includegraphics[width=0.99\linewidth]{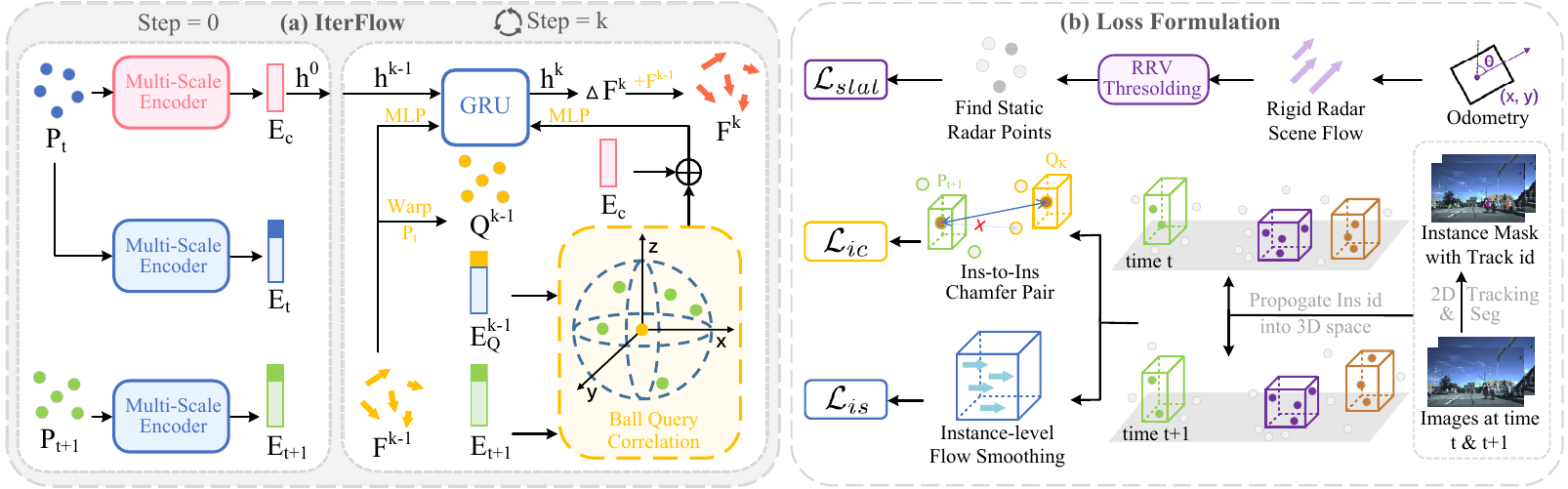}
   \caption{\textbf{Overall architecture of our proposed method.} The process of the $k$th scene flow iteration is depicted on the left and the detailed loss formulation process in the training stage is given on the right. $\oplus$ represents concatenation. }
   \label{fig:pipeline}
\end{figure*}

\noindent\textbf{Problem Definition.} 
Radar scene flow estimation aims to recover the 3D motion field between two consecutive radar scans, $\mathbf{P}_{t}\in\mathbb{R}^{N_1\times 5}$ and $\mathbf{P}_{t+1}\in\mathbb{R}^{N_2\times 5}$, where $N_1$, $N_2$ denote the number of radar points in each frame. Every radar point consists of five attributes: its 3D coordinates, radar cross-section (RCS), and relative radial velocity (RRV). The coordinates of points in the source and target frame are denoted as $x_i\in\mathbf{P}_{t}$ and $y_i\in\mathbf{P}_{t+1}$, respectively. 
Following previous label-free settings~\cite{ding2023hidden,wu2025tars,zhai2025dmrflow}, while data from other modalities may be accessible during training, only radar data is required as input during inference. 
During testing, given $\mathbf{P}_{t}$ and $\mathbf{P}_{t+1}$, the model predicts the scene flow $\mathbf{F}\in\mathbb{R}^{N_1\times 3}$ for each radar point in the source frame $\mathbf{P}_{t}$.

\noindent\textbf{Method Overview.} 
The overall architecture of our proposed method comprises an iterative scene flow estimation network, termed IterFlow (Fig.~\ref{fig:pipeline}a), and a corresponding loss formulation component (Fig.~\ref{fig:pipeline}b). 
During training, the consecutive radar point clouds $\mathbf{P}_{t}$ and $\mathbf{P}_{t+1}$ are fed into IterFlow to generate the final scene flow prediction $\mathbf{F}^{\mathrm{K}}\in\mathbb{R}^{N_1\times 3}$ after $\mathrm{K}$ steps of iterative refinement, which will be introduced in Sec.~\ref{sec:Iterflow}. Subsequently, auxiliary 2D image and odometry are used to construct three losses for optimizing the predicted flows: $\mathcal{L}_{total} = \mathcal{L}_{stat} + \mathcal{L}_{ic} + \mathcal{L}_{is}$. 
Here $\mathcal{L}_{stat}$ denotes the rigid static loss defined in Sec.~\ref{sec:stat}; $\mathcal{L}_{ic}$ and $\mathcal{L}_{is}$ correspond to our instance-aware Chamfer loss and instance-level flow smoothness loss, respectively, which will be detailed in Sec.~\ref{sc:ins_loss}.

\begin{figure}[t!]
  \centering
   \includegraphics[width=0.97\linewidth]{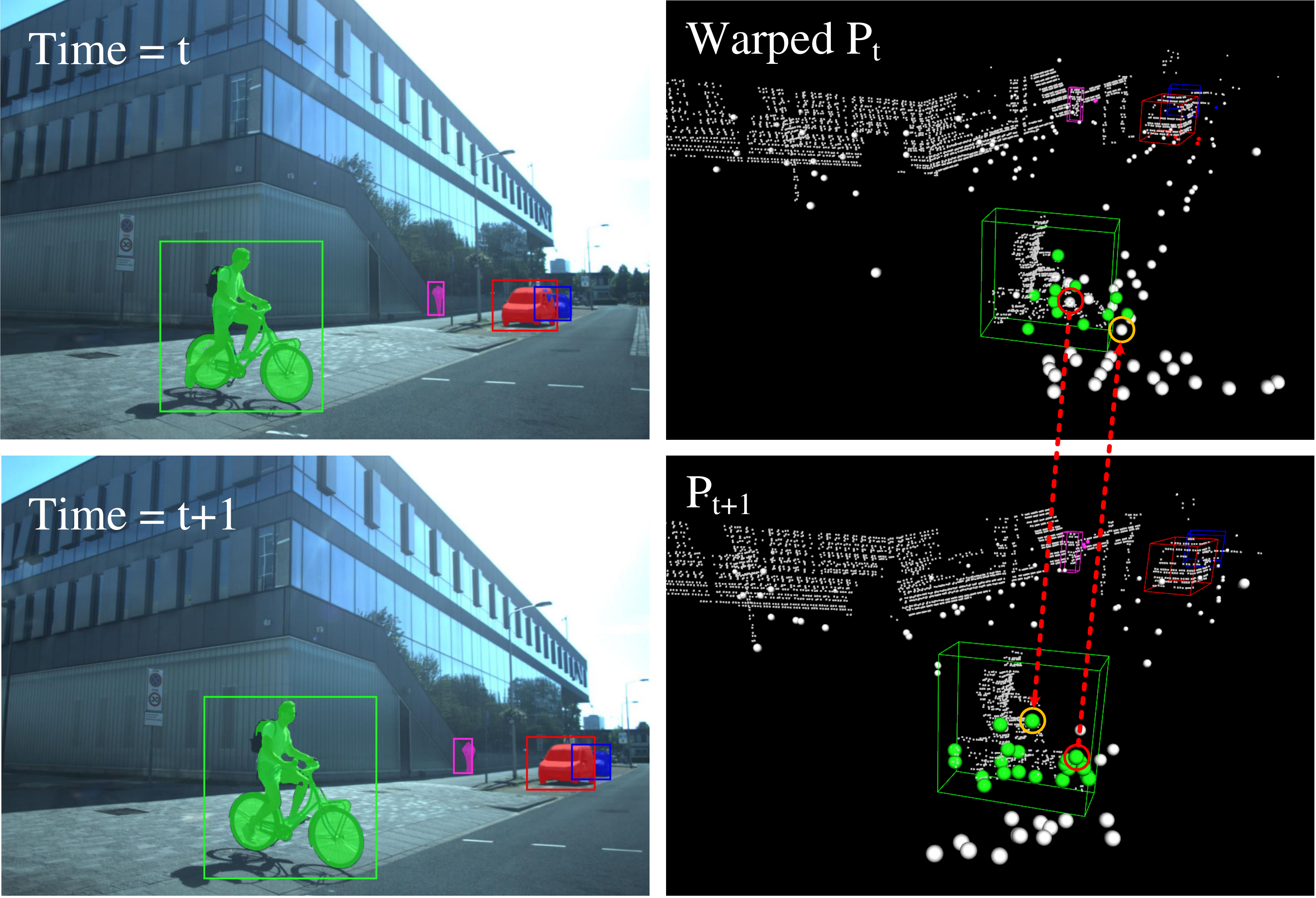}
   \caption{\textbf{An example of mismached chamfer pairs.} The top and bottom rows show two consecutive frames at time $t$ and time $t+1$, respectively. The smaller points are LiDAR points for auxiliary visualization, while the larger balls are radar points.
   $\mathbf{P}_{t}$ is first warped by estimated scene flow and then used to calculate chamfer loss with $\mathbf{P}_{t+1}$. The orange circles indicate the current selected points for normal chamfer loss calculation, while the red circles and dashed lines highlight the possible mismatch between colored dynamic foreground points and white background points when instance-level guidance is unavailable.
   }
   \label{fig:mismatch}
\end{figure}

\subsection{IterFlow}\label{sec:Iterflow}
Previous radar-based scene flow estimation methods typically adopt KNN-based grouping for cross-frame correlation, such as patch-to-patch correaltion in \cite{ding2022self,ding2023hidden,zhai2025dmrflow} and cross-attention in \cite{wu2025tars}.
However, KNN grouping often leads to mismatches and ambiguities in sparse radar point clouds. Moreover, these approaches either produce final flow predictions in a single step or rely on additional auxiliary refinement modules.
To address these limitations and achieve high-accuracy flow estimation on challenging 4D radar data, we propose IterFlow, a task-specific iterative network designed to refine scene flow estimation progressively (see Fig.~\ref{fig:pipeline}a). IterFlow leverages ball query grouping to correlate cross-frame features within a constrained spatial range and employs a GRU-based~\cite{cho2014properties} recurrent update scheme for iterative scene flow refinement.

In IterFlow, $\mathbf{P}_{t}$ and $\mathbf{P}_{t+1}$ share a multi-scale encoder~\cite{ding2022self, ding2023hidden} to extract point cloud features $\mathbf{E}_{t}\in\mathbb{R}^{N_1\times (3+C)}$ and $\mathbf{E}_{t+1}\in\mathbb{R}^{N_2\times (3+C)}$, respectively. Each pointwise feature $\varphi(x_i)\in\mathbf{E}_{t}$ and $\varphi(y_i)\in\mathbf{E}_{t+1}$ consists of the original input 3D position and the feature dimension $C$. Here $\varphi(\cdot)$ represents the multi-scale encoder.
Following~\cite{wei2021pv,wang20233d}, another multi-scale encoder processes $\mathbf{P}_{t}$ to produce contextual features $\mathbf{E}_{c}\in\mathbb{R}^{N_1\times C}$, which initialize the hidden state of the GRU unit for iterative residual flow generation.

During the $k$-th iteration, $\mathbf{P}_{t}$ is warped to $\mathbf{Q}^{k-1}$ using the previous flow estimation $\mathbf{F}^{k-1}$, replacing the 3D coordinate part of $\mathbf{E}_{t}$ with the warped coordinates $q_i\in\mathbf{Q}^{k-1}$ to form the warped features $\varphi(q_i)\in\mathbf{E}_{Q}^{k-1}$. After that, a ball query operation with a limited search radius $\mathrm{R}$ is applied to each $q_i$ to find its $\mathrm{L}$ nearest neighbors in $\mathbf{P}_{t+1}$, denoted as $\mathcal{N}_\mathrm{L}=\mathcal{N}_{\mathbf{P}_{t+1}}(q_i)_\mathrm{L}$. With set abstraction in ~\cite{qi2017pointnet,qi2017pointnet++}, the ball query-based cross-frame correlation feature is then computed as:
\begin{equation}
\textbf{c}_{i}^{k} = \max_{l}({\rm MLP}({\rm \mathop{concat}_{y_l\in\mathcal{N}_\mathrm{L}}}(\varphi(y_l)\cdot\varphi(q_i), y_l - q_i))).
\end{equation}
The correlation feature $\textbf{c}_{i}^{k}\in\textbf{C}^{k}$ is concatenated with context feature $\textbf{E}_{c}$ and fed into MLP, incorporating source-frame geometry to support flow prediction.
Meanwhile, the coarse flow vector $\textbf{F}_{k-1}$ is also encoded via an MLP.
These features are fused to form the GRU input $x_{k}$, and the hidden state is updated as follows:
\begin{align}
&z^{k} = \sigma(\text{Conv}_{\text{1d}}([h^{k-1}, x^{k}], W_z)) \\
&r^{k} = \sigma(\text{Conv}_{\text{1d}}([h^{k-1}, x^{k}], W_r)) \\
&\hat{h^{k}} = \tanh(\text{Conv}_{\text{1d}}([r^{k} \odot h^{k-1}, x^{k}], W_h)) \\
& h^{k} = (1 - z^{k}) \odot h^{k-1} + z^{k} \odot \hat{h^{k}}
\end{align}
where weight matrices $W_{z}$, $W_{r}$, and $W_{h}$ can be learned during training. The updated $h^{k}$ is fed into a convolutional flow head to produce residual flow $\Delta\mathbf{F}^{k}$, and the flow prediction is iteratively refined as:
$\mathbf{F}^{k}=\mathbf{F}^{k-1}+\Delta\mathbf{F}^{k}$.

\subsection{Instance-aware Loss Functions}
\label{sc:ins_loss}

\noindent\textbf{Pointwise Instance Label Generation\footnote{Please refer to Sec.~\ref{sec:ra_ins} in the appendix for detailed process of generating instance label for each radar point.}.} 
\label{sec:label}
As depicted in Fig.~\ref{fig:pipeline}b, the images at time $t$ and $t+1$ are fed into off-the-shelf 2D tracking~\cite{khanam2024yolov11} and pretrained segmentation models~\cite{kirillov2023segment} to generate instance segmentation masks with tracked IDs. 
Through 2D-to-3D back-projection, corresponding 2D instance label $g \in {1,2,3,\dots,G}$ or background label $g=0$ can be propogated into 3D space and assigned to each 3D radar point. Here, $G$ denotes the maximum number of foreground instances observed across the source and target frames. The retrieved pointwise 3D instance label is denoted as $\phi(\cdot)$. The radar points are then grouped by their instance labels to form instance-level point sets in both frames: $\mathbf{S}_{t}^{g}=\{x_{i}~|\phi(x_{i})=g, x_{i}\in\mathbf{P}_{t}\}$ and $\mathbf{S}_{t+1}^{g}=\{y_{j}~|\phi(y_{j})=g, y_{j}\in\mathbf{P}_{t+1}\}$.

\noindent\textbf{Instance-aware Chamfer Loss.} 
The commonly used Chamfer loss is calculated based on the mutual nearest neighbor distance between two point clouds.
However, due to the inherent sparsity and disorder of point clouds, mismatched Chamfer pairs are inevitable, and this issue becomes more pronounced with sparse 4D radar data. As illustrated in Fig.~\ref{fig:mismatch}, radar points on objects are sparse and often surrounded by static background points or noise. In such cases, the nearest neighbor of a dynamic foreground point in $P_{t+1}$ within the warped $P_{t}$ may incorrectly correspond to a background point, and vice versa. This misalignment significantly hinders the network's ability to learn accurate motion for dynamic foreground objects.

To mitigate this issue, we propose calculating the Chamfer loss exclusively between point pairs that belong to the same instance, utilizing the pointwise instance label obtained in the previous step (\textit{e.g.} green cyclist points in warped $P_{t}$ and $P_{t+1}$). With assistance from 2D semantic information, our instance-aware Chamfer loss $\mathcal{L}_{ic}$ (Eq.~\ref{eq:L_ic}) effectively reduces the negative impact of mismatched point pairs and prevents confusion between dynamic foreground and static background points. 
\begin{small} 
\begin{equation}
\mathcal{L}_{ic} = \frac{1}{G} \sum_{g=1}^{G}(\sum_{x_i\in\mathbf{S}_{t}^{g}}\frac{D(\mathbf{\hat{P}}_{t,i},\mathbf{P}_{t+1})}{|\mathbf{S}_{t}^{g}|} + \sum_{y_j\in\mathbf{S}_{t+1}^{g}}\frac{D(\mathbf{P}_{t+1,j},\mathbf{\hat{P}}_{t})}{|\mathbf{S}_{t+1}^{g}|}),
\label{eq:L_ic}
\end{equation}
\end{small}
where $\mathbf{\hat{P}}_{t}$ is the source frame point cloud $\mathbf{P}_{t}$ warped by the predicted scene flow, and $D(p,\mathbf{P})$ is the minimum distance between point $q$ and its nearest neighbor in point cloud $\mathbf{P}$:
\begin{equation}
D(q,\mathbf{P})=\min_{p_{m}\in\mathbf{P}}\lVert q-p_{m}\rVert_2^2
\end{equation}

\begin{figure}[t!]
  \centering
   \includegraphics[width=0.95\linewidth]{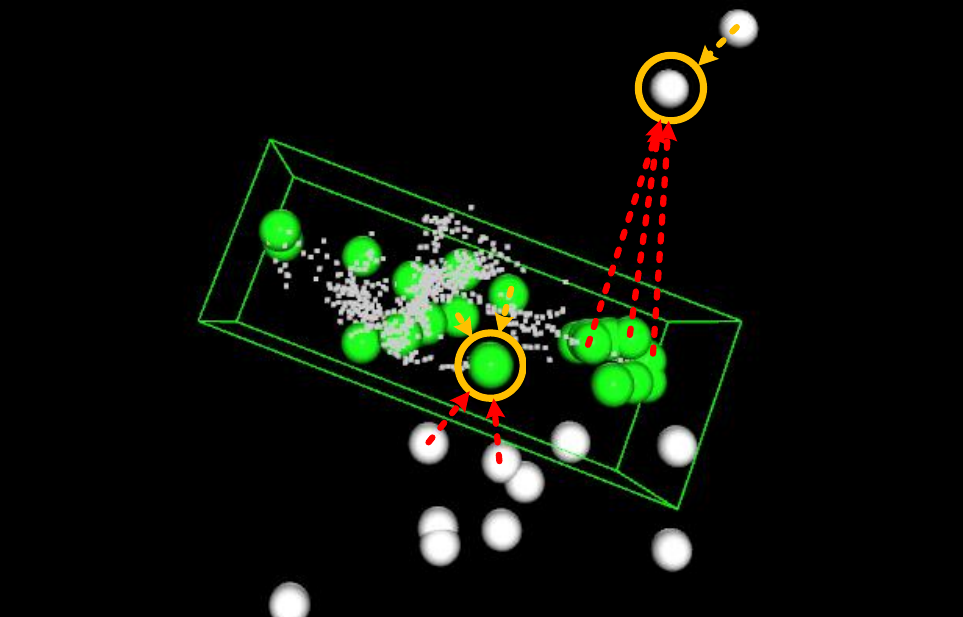}
   \caption{\textbf{An example of wrong KNN-based spatial flow smoothing.} The figure shows the bird's-eye view of a cyclist (green bounding box). Foreground radar points of the cyclist are painted in green. The orange circles highlight the current center points for KNN search, while the red dashed lines represent the confusion between foreground dynamic and background static points caused by the KNN-based flow smoothing in previous methods.}
   \label{fig:err_knn}
\end{figure}

\noindent\textbf{Instance-level Flow Smoothness Loss.} 
Previous works~\cite{ding2022self,ding2023hidden,zhai2025dmrflow,wu2025tars} typically adopt a spatial smoothness loss that enforces flow consistency within a KNN neighborhood. However, this KNN-based flow smoothing introduces potential risks of incorrect flow regularization between dynamic foreground and static background points, especially since radar point clouds are unevenly distributed and often extremely sparse in certain regions. A real-world example is shown in Fig.~\ref{fig:err_knn}, where both foreground and background points are prone to mismatches during KNN-based flow smoothing. The resulting enforced consistency between incorrect point pairs can significantly degrade network performance.

To address this problem, we introduce an instance-level flow smoothness loss $\mathcal{L}_{is}$. Unlike previous methods that apply flow consistency constraints for all radar points in the source frame, $\mathcal{L}_{is}$ focuses exclusively on maintaining flow smoothness within each foreground instance. This design effectively mitigates the influence of noisy background points and prevents mismatches between dynamic foreground and static regions. The mean predicted scene flow of each instance is calculated as: $\mathbf{F}_{mean}^g= \frac{1}{|\mathbf{S}_{t}^g|} \sum_{x_i\in\mathbf{S}_{t}^{g}}{\mathbf{F}_{i}^K}$, and $\mathcal{L}_{is}$ is formulated as:
\begin{equation}
\mathcal{L}_{is} = \frac{1}{G} \sum_{g=1}^{G}\frac{1}{|\mathbf{S}_{t}^{g}|}\sum_{x_i\in\mathbf{S}_{t}^{g}}\lVert\mathbf{F}_{i}^K-\mathbf{F}_{mean}^g\rVert_2.
\label{eq:L_ics}
\end{equation}

\begin{table*}[ht!]
\caption{\textbf{
Quantitative Evaluation and Model Complexity Comparison on VoD validation set.
} 
In the Category (Cat.) column, existing methods are classified depending on the input modality used in their original work. In the Supervision (Sup.) column, the methods are categorized as fully supervised (Full), self-supervised (Self), or cross-supervised (Cross). R represents radar point clouds input.
}
\centering
    \resizebox{0.99\textwidth}{!}{
    \large
\begin{tabular}{lcc|c|cccccc|c|c|c|c}
\toprule[2pt]
& \multicolumn{3}{l|}{} 
& \multicolumn{6}{c|}{Overall}             & \multicolumn{1}{c|}{Moving}                                          & \multicolumn{1}{c|}{Static}    &      \multicolumn{2}{c}{Model Complexity }                              \\ 
\hline
\multicolumn{1}{c|}{Cat.}              & \multicolumn{1}{c|}{Method}  & \multicolumn{1}{c|}{Inp.} & Sup.                  &  & \quad EPE [m]$\downarrow$ \quad                              & \quad AccS [\%]$\uparrow$ \quad                            & \quad AccR [\%]$\uparrow$ \quad                            & \quad \enspace RNE [m]$\downarrow$ \enspace \quad &    & \quad MRNE [m]$\downarrow$ \quad                               &  \quad SRNE [m]$\downarrow$   \quad  & Params [B]$\downarrow$     & GFLOPs$\downarrow$                      \\ 
\midrule[1pt]
\multicolumn{1}{l|}{\multirow{7}{*}{\makecell{LiDAR-\\based}}} & \multicolumn{1}{l|}{Flow4D~\cite{kim2025flow4d}}&R  & Full  &  
& $0.2555$    & $4.48$     & $19.88$    & $0.1032$    &  &  
 $0.1194$      &   $0.1007$        &$4.6$M  & -
\\
\multicolumn{1}{l|}{} & \multicolumn{1}{l|}{BiFlow~\cite{cheng2022bi}} &R  & Full  &  
& $0.2140$   & $18.14$     & $35.07$    & $0.0859$    &  &  
 $0.1235$      &   $0.0823$      &$8.2$M & $1.01$
\\
\multicolumn{1}{l|}{} & \multicolumn{1}{l|}{DeFlow~\cite{zhang2024deflow}} &R & Full  &  
& $0.2015$    & $8.94$     & $29.01$    & $0.0813$    &  &  
 $0.0981$      &   $0.0789$       &$6.9$M  & -
\\
\multicolumn{1}{l|}{} & \multicolumn{1}{l|}{PointPWC~\cite{wu2020pointpwc}} &R  & Full  & 
& $0.1271$    & $23.80$    & $53.00$    & $0.0512$    &  &  
 $0.0859$      &   $0.0478$      &$7.7$M  & $1.26$
\\
\multicolumn{1}{l|}{} & \multicolumn{1}{l|}{PV-RAFT~\cite{wei2021pv}} &R  & Full  & 
& $0.1140$    & $23.24$     & $58.07$    & $0.0459$    &  &  
 $0.0861$    &    $0.0418$       &$192$K  & $0.63$
\\
\multicolumn{1}{l|}{} & \multicolumn{1}{l|}{PointPWC~\cite{wu2020pointpwc}} &R  & Self  &  
& $0.4314$     & $0.02$      & $0.17$   & $0.1735$    &  &  
 $0.1703$      &   $0.1731$      &$7.7$M  & $1.26$
\\
\multicolumn{1}{l|}{} & \multicolumn{1}{l|}{FlowStep3D~\cite{kittenplon2021flowstep3d}}  &R  & Self  &  
& $0.2607$     & $2.54$      & $14.62$   & $0.1050$    &  &  
 $0.1261$      &   $0.1025$      &$689$K  & $0.85$
\\
\midrule[1pt]
\multicolumn{1}{l|}{\multirow{3}{*}{\makecell{Radar-\\based}}}   &\multicolumn{1}{l|}{RaFlow~\cite{ding2022self}}  &R  & Self  &  
& $0.2753$     & $3.49$      & $17.48$    & $0.1105$    &  &   $0.1372$      &   $0.1081$    &$4.1$M    &$12.78$
\\
\multicolumn{1}{l|}{} & \multicolumn{1}{l|}{CMFlow~\cite{ding2023hidden}} &R  & Cross  &  
& $0.1600$     & $19.83$      & $42.38$    & $0.0643$   &  &  
 $0.0838$      &   $0.0625$       &$4.2$M  & $12.83$
\\
\multicolumn{1}{l|}{} & \multicolumn{1}{l|}{IterFlow (ours)}  &R & Cross  &  
& $\textbf{0.1045}$    & $\textbf{33.40}$      & $\textbf{63.75}$     & $\textbf{0.0420}$    &  &  
 $\textbf{0.0833}$      &   $\textbf{0.0380}$    &  $\textbf{113K}$  & $\textbf{0.40}$ 
\\
\bottomrule[2pt]
\end{tabular}
}
\label{tab:main}
\end{table*}

\begin{table*}[h!]
\caption{\textbf{Quantitative Evaluation on Network Architecture and Loss Scalability on VoD validation set. 
} $\mathcal{L}_{c}$, $\mathcal{L}_{s}$, $\mathcal{L}_{g}$ denote the official chamfer loss, local flow smoothness loss and local geometric consistency loss from PointPWC~\cite{wu2020pointpwc}. $\mathcal{L}_{rd}$, $\mathcal{L}_{sc}$, $\mathcal{L}_{ss}$ represent the radial displacement loss, soft chamfer loss and spatial smoothness loss from RaFlow~\cite{ding2022self}. }
\centering
    \resizebox{0.99\textwidth}{!}{
    \large
\begin{tabular}{rlc|cccccc|ccc|ccc}
\toprule[2pt]
& & \multicolumn{1}{l|}{} 
& \multicolumn{6}{c|}{Overall}             & \multicolumn{3}{c|}{Moving}                                          & \multicolumn{3}{c}{Static}                                           \\ 
\hline
& \multicolumn{1}{l|}{Method}             & Loss                 &  & \quad EPE [m]$\downarrow$ \quad                              & \quad AccS [\%]$\uparrow$ \quad                            & \quad AccR [\%]$\uparrow$                           & \quad \enspace RNE [m]$\downarrow$ \enspace  &  &  &  MRNE [m]$\downarrow$                              &  &  & SRNE [m]$\downarrow$                              &  \\ 
\midrule[1pt]
1) &\multicolumn{1}{l|}{PointPWC~\cite{wu2020pointpwc}} &\cite{wu2020pointpwc}: $\mathcal{L}_{c}+\mathcal{L}_{s}+\mathcal{L}_{g}$  &  
& $0.4314$     & $0.02$      & $0.17$   & $0.1735$    &  &  
& $0.1703$    &  &  & $0.1731$     &  
\\
\rowcolor{gray!20}
2)& \multicolumn{1}{l|}{PointPWC~\cite{wu2020pointpwc}} &Ours: $\mathcal{L}_{stat}+\mathcal{L}_{ic}+\mathcal{L}_{is}$  &  
& $\textbf{0.1228}$     & $\textbf{25.07}$      & $\textbf{54.45}$   & $\textbf{0.0495}$    &  &  
& $\textbf{0.0875}$    &  &  & $\textbf{0.0457}$   &  
\\
\midrule[1pt]
3) &\multicolumn{1}{l|}{RaFlow~\cite{ding2022self}}   &~\cite{ding2022self}: $\mathcal{L}_{rd}+\mathcal{L}_{sc}+\mathcal{L}_{ss}$  &  
& $0.2753$     & $3.49$      & $17.48$    & $0.1105$    &  &  
& $0.1372$    &  &  & $0.1081$     &  
\\
\rowcolor{gray!20}
4) &
\multicolumn{1}{l|}{RaFlow~\cite{ding2022self}}   &Ours: $\mathcal{L}_{stat}+\mathcal{L}_{ic}+\mathcal{L}_{is}$  &  
& $\textbf{0.1163}$     & $\textbf{28.90}$      & $\textbf{58.30}$   &$\textbf{0.0468}$    &  &  
& $\textbf{0.0917}$    &  &  &$\textbf{ 0.0426}$     &  
\\
\midrule[1pt]
5)
&\multicolumn{1}{l|}{IterFlow}   &\cite{wu2020pointpwc}: $\mathcal{L}_{c}+\mathcal{L}_{s}+\mathcal{L}_{g}$  & 
& $0.4084$     & $2.13$     & $12.05$    & $0.1638$    &  &  
& $0.1280$   &  &  & $0.1669$     &  
\\
6)& \multicolumn{1}{l|}{IterFlow}   &\cite{ding2022self}: $\mathcal{L}_{rd}+\mathcal{L}_{sc}+\mathcal{L}_{ss}$   &  
& $0.1872$     & $19.53$      & $41.44$    & $0.0751$    &  &  
& $0.0851$    &  &  & $0.0740$     &  
\\
\rowcolor{gray!20}
7)& \multicolumn{1}{l|}{IterFlow}  &Ours: $\mathcal{L}_{stat}+\mathcal{L}_{ic}+\mathcal{L}_{is}$  &  
& $\textbf{0.1045}$    & $\textbf{33.40}$      & $\textbf{63.75}$     & $\textbf{0.0420}$    &  &  
& $\textbf{0.0833}$    &  &  & $\textbf{0.0380}$     &  
\\
\bottomrule[2pt]
\end{tabular}
}
\label{tab:VOD}
\end{table*}

\subsection{Rigid Static Loss}
\label{sec:stat}
The rigid static loss $\mathcal{L}_{stat}$ measures the difference between the estimated scene flow of static points and their corresponding rigid flow based on the ground-truth vehicle odometry.
Following previous methods~\cite{wu2025tars,zhai2025dmrflow}, we first compensate radar RRV with known ego motion to obtain absolute radial velocity (ARV) for each point in $\mathbf{P}_{t}$. 
After that, static radar points can be identified by ARV thresholding and merged into a static point set $\mathbf{S}_{r}=\{x_{i}~|ARV_{i}<\gamma_{thre}, x_{i}\in\mathbf{P}_{t}\}$. 
Then $\mathcal{L}_{stat}$ can be formulated as Eq.~\ref{eq:L_iter}, where $\mathbf{T}$ is the ground-truth ego-motion transformation.
\begin{equation}
\label{eq:L_iter}
\mathcal{L}_{stat} = \frac{1}{|\mathbf{S}_{r}|} \sum_{x_i\in\mathbf{S}_{r}}{\lVert{(\mathbf{F}_{i}^{\mathrm{K}} -(\mathbf{T}\circ x_{i}- x_{i}))}\rVert_2},
\end{equation}

\section{Experiments}

\noindent\textbf{Datasets.} 
Following~\cite{ding2023hidden,zhai2025dmrflow, wu2025tars}, we conduct experiments on the real-world View-of-Delft (VoD) dataset~\cite{palffy2022multi}, which provides 8,600 frames of synchronized and calibrated odometry, 3D LiDAR point clouds, 4D radar data and RGB images. The average ratio
between radar and LiDAR resolution is 2.5 on the VoD dataset.
The radar points within the viewing frustum are taken as input, which are restricted in a height range of $[-3\mathrm{m},3\mathrm{m}]$. 
Since the VoD dataset does not provide ready-made scene flow ground truth, we adopt the commonly used preprocessing methods to generate scene flow labels from annotated 3D tracking boxes for the training set (5,066 frames) and validation set (1,273 frames) according to the official split\footnote{Please refer to Sec.~\ref{sec:vod} and Sec.~\ref{sec:rigid_sf} in appendix for detailed introduction of VoD dataset and process of annotating rigid scene flow based on 3D tracking boxes.}.

\noindent\textbf{Implementation Details.}
\label{sec:imp}
Following the previous radar-based methods~\cite{ding2022self,ding2023hidden}, $N_1=N_2=256$ radar points are randomly sampled in each frame as input and the ARV threshold $\gamma_{thre}$ is set to $0.1\mathrm{m}/s$ to identify static points. 
For the ball query correlation, we sample $\mathrm{L}=8$ neighbors with a limited search radius of $\mathrm{R}=1\mathrm{m}$. Scene flow is updated for $\mathrm{K}=12$ iterations in both training and evaluation stage\footnote{Please refer to Sec.~\ref{sec:abla_iter} in the appendix for hyperparameter sensitivity analysis of $\mathrm{L}, \mathrm{R}$ and $\mathrm{K}$.}. In order to generate reliable instance labels for radar points from 2D images, we adopt the officially released YOLO11-l~\cite{khanam2024yolov11} model and the huge version of Segment Anything Model (SAM)~\cite{kirillov2023segment} for 2D tracking and segmentation. IterFlow is implemented in PyTorch~\cite{paszke2019pytorch} and trained for 150 epochs with a batch size of 8. The Adam optimizer with an initial learning rate of 0.001 is used. For a fair comparison with the baselines, we use their official loss configuration and hyperparameter settings for network retraining on the VoD radar scene flow dataset. In addition, for the cross-modal supervised CMFlow~\cite{ding2023hidden}, we generate extra required optical flow labels and pseudo scene flow labels by adopting their officially released  pretrained 2D optical flow estimation model and 3D tracking results.

\noindent\textbf{Evaluation Metrics.}
Following~\cite{ding2022self,ding2023hidden,zhai2025dmrflow,wu2025tars}, we use standard metrics for evaluation: 1) \underline{EPE}: average 3D end-point-error measures the L2 distance between
ground truth flow vectors and scene flow predictions; 2) \underline{AccS}/\underline{AccR}: the ratio of points that meet a strict/relaxed condition, \textit{i.e.} $\mathrm{EPE}<0.05/0.1\mathrm{m}$ or the relative error $\mathrm{<}5\%/10\%$; 3) \underline{RNE}: resolution-normalized EPE by dividing EPE by the ratio of 4D radar and LiDAR resolution, to accommodate radar sensors with different resolution. 4) \underline{MRNE}/\underline{SRNE}: RNE for moving points and static points, respectively.

\begin{figure*}[ht!]
  \centering
   \includegraphics[width=0.97\linewidth]{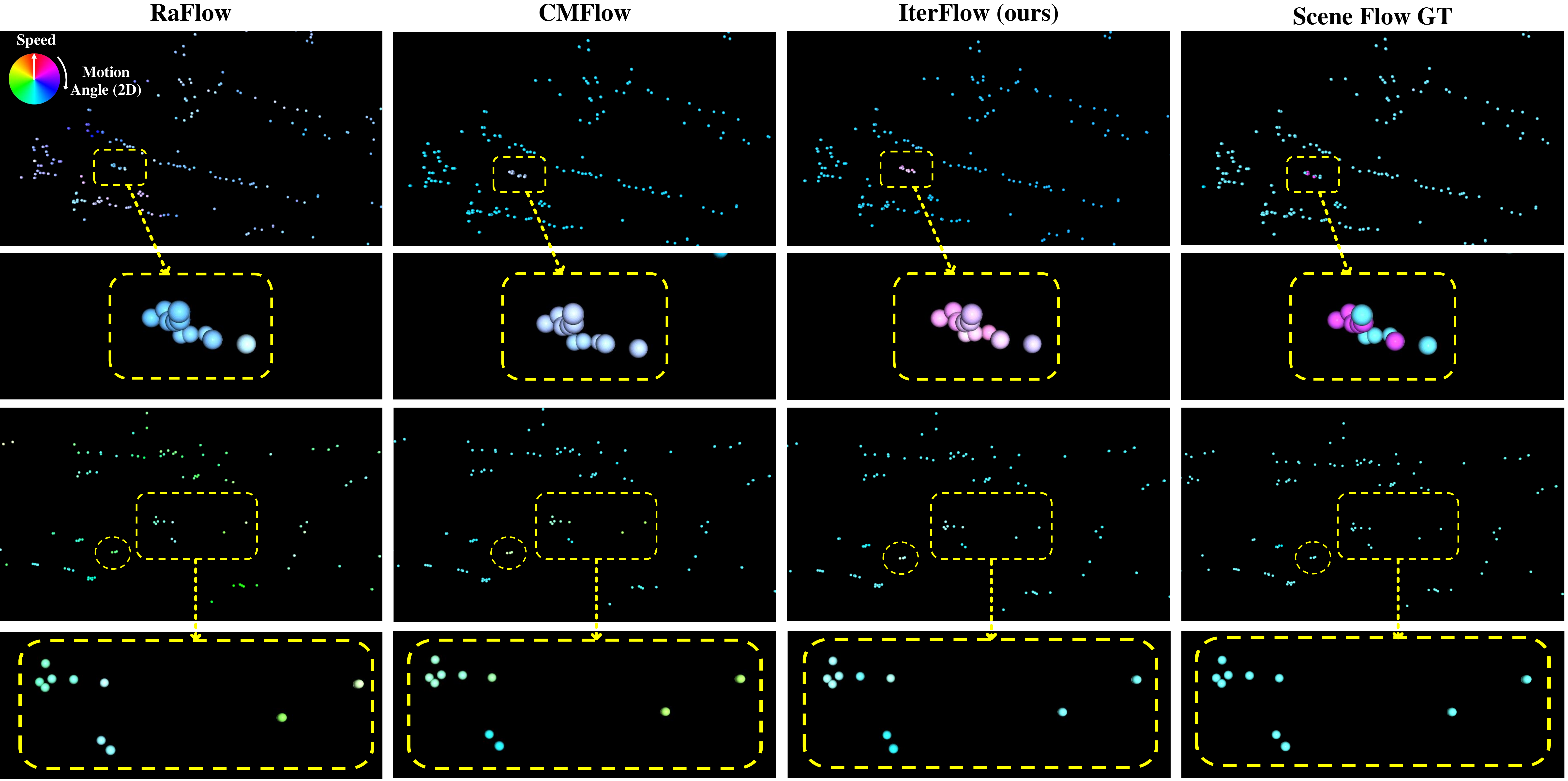}
   \caption{\textbf{Qualitative Results on VoD validation dataset}. The first row and the third row display two separate traffic scenes; while the second row and fourth row show the zoomed-in results of the regions highlighted with yellow rectangles. As shown in the legend, the direction and magnitude of scene flow vectors are employed as hue and saturation, respectively.}
   \label{fig:vis}
\end{figure*}

\subsection{Main Results}
\noindent\textbf{Overall Quantitative Comparison.}
We compare our method with state-of-the-art scene flow estimation models on the VoD dataset and the experimental results are summarized in Table~\ref{tab:main}. Note that fully-supervised methods are trained with the radar scene flow ground truth derived from the annotated 3D tracking boxes provided by the dataset, and CMFlow~\cite{ding2023hidden} is trained with cross-modal supervisions including pseudo scene flow generated based on the 3D tracking results from a pretrained multi-object tracking model~\cite{weng20203d}; both are supervised by direct 3D motion cues, while the remaining methods (including ours) operate without access to ground-truth tracking boxes or dense LiDAR point clouds.

Compared to the previous best radar-based model, CMFlow~\cite{ding2023hidden}, which employs a complex multi-task framework and uses a total of seven loss terms for training (including three self-supervised losses and four cross-modal losses), our task-specific IterFlow is trained with only three losses and achieves superior performance across all evaluation metrics. In particular, IterFlow yields a 34.7\% performance improvement on the EPE metric, while increasing AccS and AccR by 13.6\% and 21.4\%, respectively. IterFlow also brings a 39.2\% error reduction on the SRNE metric and achieves the best performance on the MRNE metric, generating more accurate flow estimation in static regions while maintaining strong performance in dynamic areas. This demonstrates that effectively leveraging instance-aware guidance from 2D images allows our weakly supervised cross-modal learning framework to substantially outperform CMFlow, which relies on pseudo 3D scene flow and 2D optical flow labels for supervision.

When compared to LiDAR-based approaches, our method even surpasses the best fully supervised model, PV-RAFT~\cite{wei2021pv}, achieving performance improvements across all metrics, enhancing the prediction accuracy on the EPE metric by 8.3\%. This result highlights that our ball query–based correlation operation is more robust in sparse radar scenarios than the KNN-based and voxel-based correlation modules used in PV-RAFT.
Overall, our approach establishes a new state of the art on the VoD radar scene flow dataset, validating the effectiveness of our network architecture and loss design.

\begin{table*}[ht!]
\caption{\textbf{Ablation Study on Loss Terms on VoD validation set.} $\mathcal{L}_{sc}$ is the soft chamfer loss without instance-aware guidance and $\mathcal{L}_{ss}$ is the KNN-based spatial smoothness loss; both are from RaFlow~\cite{ding2022self}. 
$\star$ denotes using the pointwise instance labels from annotated 3D tracking boxes in VoD dataset for instance-aware loss calculation.
}
\centering
    \resizebox{0.99\textwidth}{!}{
    \large
\begin{tabular}{l|cccccc|ccc|ccc}
\toprule[2pt]
 \multicolumn{1}{c|}{} 
& \multicolumn{6}{c|}{Overall}             & \multicolumn{3}{c|}{Moving}                                          & \multicolumn{3}{c}{Static}                                           \\ 
\hline
 Loss                 &  & \quad EPE [m]$\downarrow$ \quad                              & \quad AccS [\%]$\uparrow$ \quad                            & \quad AccR [\%]$\uparrow$ \quad                            & \quad \enspace RNE [m]$\downarrow$ \enspace  &  &  &  MRNE [m]$\downarrow$                            &  &  & SRNE [m]$\downarrow$                             &  \\ 
\midrule[1pt]
\lfbox[rounded,border-radius=3pt,border-width=0pt,padding=2pt,margin=0pt,background-color=mypurple!15]{$\mathcal{L}_{stat}$}
$+$
\lfbox[rounded,border-radius=3pt,border-width=0pt,padding=2pt,margin=0pt,background-color=white]{$\mathcal{L}_{sc}$}
 
&  
& $0.1512$     & $23.47$     & $51.11$   & $0.0609$    &  &  
& $0.1021$   &  &  & $0.0571$     &  
\\
\lfbox[rounded,border-radius=3pt,border-width=0pt,padding=2pt,margin=0pt,background-color=mypurple!15]{$\mathcal{L}_{stat}$} 
$+$
\lfbox[rounded,border-radius=3pt,border-width=0pt,padding=2pt,margin=0pt,background-color=yellow!30]{$\mathcal{L}_{ic}$} 
 &  
& $0.1249$     & $28.83$      & $58.28$    & $0.0503$    &  &  
& $0.0937$    &  &  & $0.0461$     &  
\\
\lfbox[rounded,border-radius=3pt,border-width=0pt,padding=2pt,margin=0pt,background-color=mypurple!15]{$\mathcal{L}_{stat}$} 
$+$
\lfbox[rounded,border-radius=3pt,border-width=0pt,padding=2pt,margin=0pt,background-color=yellow!30]{$\mathcal{L}_{ic}$} 
$+$
\lfbox[rounded,border-radius=3pt,border-width=0pt,padding=2pt,margin=0pt,background-color=white]{$\mathcal{L}_{ss}$} 
&  
& $0.1191$     & $28.76$     & $57.43$    & $0.0479$   &  &  
& $0.0778$   &  &  & $0.0447$     &  
\\
\lfbox[rounded,border-radius=3pt,border-width=0pt,padding=2pt,margin=0pt,background-color=mypurple!15]{$\mathcal{L}_{stat}$} 
$+$
\lfbox[rounded,border-radius=3pt,border-width=0pt,padding=2pt,margin=0pt,background-color=yellow!30]{$\mathcal{L}_{ic}$} 
$+$
\lfbox[rounded,border-radius=3pt,border-width=0pt,padding=2pt,margin=0pt,background-color=myblue!15]{$\mathcal{L}_{is}$} 

&  
& $0.1045$    & $33.40$      & $63.75$     & $0.0420$    &  &  
& $0.0833$    &  &  & $\textbf{0.0380}$     &  
\\

\lfbox[rounded,border-radius=3pt,border-width=0pt,padding=2pt,margin=0pt,background-color=mypurple!15]{$\mathcal{L}_{stat}$} 
$+$
\lfbox[rounded,border-radius=3pt,border-width=0pt,padding=2pt,margin=0pt,background-color=yellow!30]{$\mathcal{L}_{ic}^{\star}$} 
$+$
\lfbox[rounded,border-radius=3pt,border-width=0pt,padding=2pt,margin=0pt,background-color=myblue!15]{$\mathcal{L}_{is}^{\star}$} 

&  
& $\textbf{0.1041}$    & $\textbf{34.12}$      & $\textbf{64.42}$     & $\textbf{0.0419}$    &  &  
& $\textbf{0.0740}$    &  &  & $0.0385$     &  
\\
\bottomrule[2pt]
\end{tabular}
}
\label{tab:loss}
\end{table*}

\begin{table*}[ht!]
\caption{\textbf{Ablation Study on Grouping Method for Cross-frame Correlation on VoD validation set.}
}
\centering
    \resizebox{0.99\textwidth}{!}{
    \large
\begin{tabular}{l|cccccc|ccc|ccc}
\toprule[2pt]
 \multicolumn{1}{c|}{} 
& \multicolumn{6}{c|}{Overall}             & \multicolumn{3}{c|}{Moving}                                          & \multicolumn{3}{c}{Static}                                           \\ 
\hline
Grouping Method                &  & \quad EPE [m]$\downarrow$ \quad                              & \quad AccS [\%]$\uparrow$ \quad                            & \quad AccR [\%]$\uparrow$ \quad                            & \quad \enspace RNE [m]$\downarrow$ \enspace  &  &  &  MRNE [m]$\downarrow$                            &  &  & SRNE [m]$\downarrow$                             &  \\ 
\midrule[1pt]
KNN
&  
& $0.1817$     & $19.68$     & $40.98$   & $0.0729$    &  &  
& $0.1028$   &  &  & $0.0703$     &  
\\
Ball Query
&  
& $\textbf{0.1045}$    & $\textbf{33.40}$      & $\textbf{63.75}$     & $\textbf{0.0420}$    &  &  
& $\textbf{0.0833}$    &  &  & $\textbf{0.0380}$     &  
\\
\bottomrule[2pt]
\end{tabular}
}
\label{tab:group_ablation}
\end{table*}

\noindent\textbf{Model Size and Computational Cost Comparison.} 
As shown in the rightmost two columns in Table~\ref{tab:main}, our proposed IterFlow is computational efficient and requires the minimum parameter size among all scene flow estimation methods. Compared to the best radar-based CMFlow~\cite{ding2023hidden}, IterFlow has $\sim40\times$ fewer parameters and $\sim30\times$ lower GFLOPs while achieving better performance. It illustrates that our task-specific network design is more lightweight and effective.
In particular, PV-RAFT~\cite{wei2021pv} and FlowStep3D~\cite{kittenplon2021flowstep3d} are also iterative, but these methods require more parameters than ours. 
This is because IterFlow adopts a more concise yet effective ball query-based cross-frame correlation design, rather than the original KNN-based and multi-scale voxel-based correlation branches in PV-RAFT and global correlation in FlowStep3D.

\noindent\textbf{Experiments on IterFlow and Loss Scalability.}
To further validate the effectiveness of our IterFlow network and loss designs, we conduct cross-comparative experiments between different architectures and loss combinations. Specifically, we adopt self-supervised loss functions from LiDAR-based PointPWC~\cite{wu2020pointpwc} and radar-based RaFlow~\cite{ding2022self} to train IterFlow, and conversely apply our proposed loss functions to train the original PointPWC and RaFlow networks. 

As shown in the last section of Table~\ref{tab:VOD}, when replacing the backbone of PointPWC and RaFlow with IterFlow while keeping their respective loss combinations (rows 5 and 6), the models exhibit consistently superior performance compared to the original versions (rows 1 and 3). Notably, when substituting the single-inference RaFlow (row 3) with our iterative IterFlow architecture (row 6), we observe a substantial improvement, with an EPE reduction of 33.6\%, demonstrating the clear advantage of our network design.

Furthermore, when replacing the original loss combinations in~\cite{wu2020pointpwc} and~\cite{ding2022self} with our proposed ones (rows 2 and 4), both networks achieve notable performance gains across all evaluation metrics. In particular, the EPE decreases by 71.5\% for PointPWC (row 2 \textit{vs} row 1) and 57.8\% for RaFlow (row 4 \textit{vs} row 3), showing that our loss functions can effectively generalize to other architectures and substantially enhance scene flow prediction accuracy.

Finally, comparing rows 5 and 7, we find that using LiDAR-based losses in \cite{wu2020pointpwc} to train our IterFlow results in dramatic performance degradation on all evaluation metrics,
highlighting that generic LiDAR-oriented self-supervised losses do not work well for radar-based scenarios. This analysis underscores the necessity and effectiveness of our loss designs for radar scene flow learning.

\noindent\textbf{Qualitative Results.}
In Fig.~\ref{fig:vis}, we compare the qualitative results of our IterFlow with the state-of-the-art radar-based scene flow estimation methods, with the ground truth scene flow serving as a reference. In general, our method performs much better than RaFlow\cite{ding2022self}, generating reliable scene flow predictions that best match the ground truth. Although CMFlow~\cite{ding2023hidden} has also demonstrated good performance, our IterFlow performs better in some difficult sparse areas, as highlighted with yellow rectangles.

\subsection{Ablation Studies}

\noindent\textbf{Ablation Study on Loss Terms.} 
Various ablation studies are conducted on IterFlow to investigate the advantage of each loss term. First, while keeping the rigid static loss $\mathcal{L}_{stat}$ unchanged, we replace our instance-aware chamfer loss $\mathcal{L}_{ic}$ with the soft chamfer loss $\mathcal{L}_{sc}$ from RaFlow\cite{ding2022self} to train the network.
As shown in Table~\ref{tab:loss}, all evaluation metrics decrease after replacing $\mathcal{L}_{ic}$, with overall performance on the EPE metric compromised by 21.1\%. The advantage of $\mathcal{L}_{ic}$ over $\mathcal{L}_{sc}$ is twofold: on one hand, $\mathcal{L}_{ic}$ only calculates the chamfer distance between points within the same instance across frames, which effectively reduces confusion caused by mismatched point pairs; on the other hand, $\mathcal{L}_{ic}$ is more robust to noise because the chamfer loss of points that cannot be successfully matched with any instance is not included in $\mathcal{L}_{ic}$.

Second, we examine the effectiveness of $\mathcal{L}_{is}$ by removing it from total loss. The experimental results in Table~\ref{tab:loss} illustrate that the addition of $\mathcal{L}_{is}$ successfully improves the prediction accuracy in both dynamic and static areas in the scene, achieving a performance improvement of 16.3\% on the overall EPE metric. 
As shown in Table~\ref{tab:loss}, directly replacing $\mathcal{L}_{ss}$ with $\mathcal{L}_{is}$ leads to a slight decrease in performance on moving objects, because dynamic foreground points only occupy a very small fraction of the scene and are more susceptible to 2D instance-level semantic accuracy. Moreover, it can be observed that with instance labels from 3D tracking ground truth, $\mathcal{L}_{ic}^{\star}$ and $\mathcal{L}_{is}^{\star}$ effectively promote the scene flow prediction performance in the moving areas.
In general, our $\mathcal{L}_{is}$ is more robust than KNN-based $\mathcal{L}_{ss}$, because $\mathcal{L}_{is}$ naturally avoids forced flow smoothing between dynamic foreground and static background points in sparse radar point clouds. This analysis demonstrates that establishing image-guided instance-level constraints promotes a more comprehensive understanding of the dynamic environment and powerfully contributes to scene flow prediction.

\noindent\textbf{Ablation Study on Grouping Method.} 
To verify the advantage of ball query-based grouping over commonly used KNN grouping for cross-frame correlation in sparse radar scenes, we replace the original ball query-based grouping in IterFlow with naive KNN grouping. As shown in Table~\ref{tab:group_ablation}, ball query-based grouping proves to be more robust in sparse radar point clouds, which consistently outperforms KNN grouping and improves the network performance on overall EPE metric by 42.5\%. This is because KNN always returns exactly $L$ neighbors based on distance rank, even when some of them are still spatially far away. In contrast, ball query first enforces spatial locality and then samples up to $L$ neighbors within a radius $R$, so different groups may contain different numbers of points. In sparse radar regions, this prevents cross-frame correspondences between points that are close only in rank but not in absolute distance, thereby reducing mismatches and improving robustness.

\noindent\textbf{Dependence Analysis of 2D Tracking and Segmentation} 

Generating pointwise instance labels through off-the-shelf camera-based models is an important component to our weakly-supervised cross-modal framework. 
As shown in Table~\ref{tab:loss}, IterFlow trained with $\mathcal{L}_{ic}$ and $\mathcal{L}_{is}$ has already performed nearly as $\mathcal{L}_{ic}^{\star}$ and $\mathcal{L}_{is}^{\star}$ on overall metrics. This demonstrates that, by effectively leveraging semantic information from images, our proposed image-based instance-aware loss functions can promisingly substitute for the requirement of 3D tracking labels. More experimental results with various combinations of 2D pipelines are given in Sec.~\ref{sec:sup_2d} in the appendix.

\section{Conclusion}

In this paper, we address 4D radar scene flow estimation through a new setting, where only RGB images and odometry avaliable for auxiliary supervision during training.
We introduce a novel framework that breaks from the prevailing trend of increasing network and loss complexity in existing methods. Specifically, we propose IterFlow, a lightweight yet effective network trained by only three targeted losses, including two instance-aware terms guided by 2D tracking and segmentation results, and one rigid static term based on odometry. Without any 3D motion supervision, our method surpasses cross-modal supervised and even fully supervised models, 
highlighting the potential of leveraging 2D semantic guidance for 4D radar scene flow estimation.

\section*{Acknowledgements}
This work was conducted during Jingyun's visit to the IMPL
Lab at SUTD, funded by China Scholarship Council. This research was supported by the Ministry of Education, Singapore, under its MOE Academic Research Fund Tier 2 (MOE-T2EP20124-0013). This work was also supported by the Key Project of Natural Science Foundation of Zhejiang Province under Grant LZ26F010003, the Key Research $\&$ Development Plan of Zhejiang Province under Grant No.2024C01010, 2024C01017, 2025C01039, the Joint $R\&D$ Program of the Yangtze River Delta Community of Sci-Tech Innovation with grant number 2024CSJGG01000.

\section*{Impact Statement}


This paper presents work whose goal is to advance the field of Machine
Learning. There are many potential societal consequences of our work, none which we feel must be specifically highlighted here.



\bibliography{example_paper}
\bibliographystyle{icml2026}

\newpage
\appendix

\section*{Appendix} 
\addcontentsline{toc}{section}{Appendix}

This appendix provides more detailed descriptions for input data and data preprocessing in Sec.~\ref{sc:data}. We also show the additional details of our method in Sec.~\ref{sec:sup_method}. Additional experimental results are presented in Sec.~\ref{sec:exp} to demonstrate the effectiveness of our method. 
The discussion of failure cases and limitation analysis is provided in Sec.~\ref{sec:sup_vis}.

\section{Input Data and Data Preprocessing Details}
\label{sc:data}

\subsection{Details of VoD dataset}
\label{sec:vod}
The View-of-Delft (VoD) dataset~\cite{palffy2022multi} is collected with a Toyota
Prius 2013 platform, which is equipped with a 3D LiDAR, a stereo camera, and a 4D ZF FRGen21 radar in complex urban traffic environments under normal weather conditions. With an annotation frequency of 10 Hz, the VoD dataset consists of 123,106 annotated 3D bounding boxes for both moving and static objects, including 26,587 pedestrians, 10,800 cyclists and 26,949 cars. While the LiDAR point clouds cover a 360 degree range, the tracking annotation and radar data are only available within the viewing frustum. For radar data, the VoD dataset provides three versions of 4D radar point clouds: single-scan, three-scan, and five-scan. For a fair comparison with previous methods~\cite{ding2022self,ding2023hidden,zhai2025dmrflow,wu2025tars}, we select the single-scan version as input, with an average of 247 radar points per frame. Moreover, every raw radar point in the VoD dataset contains seven aspects of information: 3D location $(x, y, z)$, reflectivity (RCS), relative radial velocity $v_{r}$, absolute radial velocity $v_{rc}$ and scan id $\tau$. We only use the information from the first five dimensions of each radar point as network input: 
\begin{equation}
[x, y, z, \mathrm{RCS},  v_{r}].
\end{equation}

\subsection{Rigid Scene Flow Annotations from Labeled Tracking Boxes}
\label{sec:rigid_sf}
Since pointwise scene flow annotation is impractical on large-scale real-world datasets, generating scene flow labels from annotated 3D tracking boxes is commonly used in previous studies
~\cite{jund2021scalable,zhang2024deflow, kim2025flow4d, khoche2025ssf, luo2025mambaflow,zhang2024seflow,ding2023hidden,zhai2025dmrflow, wu2025tars}.
Given two annotated tracking boxes with the same track ID in two consecutive frames $t_{0}$ and $t_{1}$, the homogeneous transformation $\mathbf{T}$ between these two bounding boxes can be directly inferred. The time interval between the two frames is represented as $\Delta t = t_{1}-t_{0}$. For a point $\mathbf{P}$ in the tracking box at $t_{0}$, its rigid scene flow $\mathbf{F}_{r}$ is calculated as follows:
\begin{equation}
   \mathbf{F}_{r}=\frac{1}{\Delta t} (\mathbf{T}\otimes\mathbf{P}-\mathbf{P}),
\end{equation}
where $\otimes$ denotes matrix multiplication. Note that the produced rigid scene flow vector is an approximation of the actual scene flow ground truth, which may not be accurate for non-rigid objects. However, this non-rigid deformation is minimized with a high frame rate.

\section{Supplementary Details of Our Method}
\label{sec:sup_method}

\subsection{Details of Instance Label Generation for Radar Points}
\label{sec:ra_ins}
Firstly, the images at time $t$ and $t+1$ are fed into a pretrained off-the-shelf 2D multi-object tracking network~\cite{khanam2024yolov11, gallagher2025surveying} to generate tracked 2D bounding boxes. Inspired by recent advances in Vision Foundation Models (VFMs)~\cite{zou2023segment, kirillov2023segment, ravi2024sam}, the 2D tracking results can be further refined into more detailed instance-level masks that are aligned with object boundaries. In this way, final instance segmentation masks with track ID can be generated.

\begin{figure}[h!]
  \centering
   \includegraphics[width=0.85\linewidth]{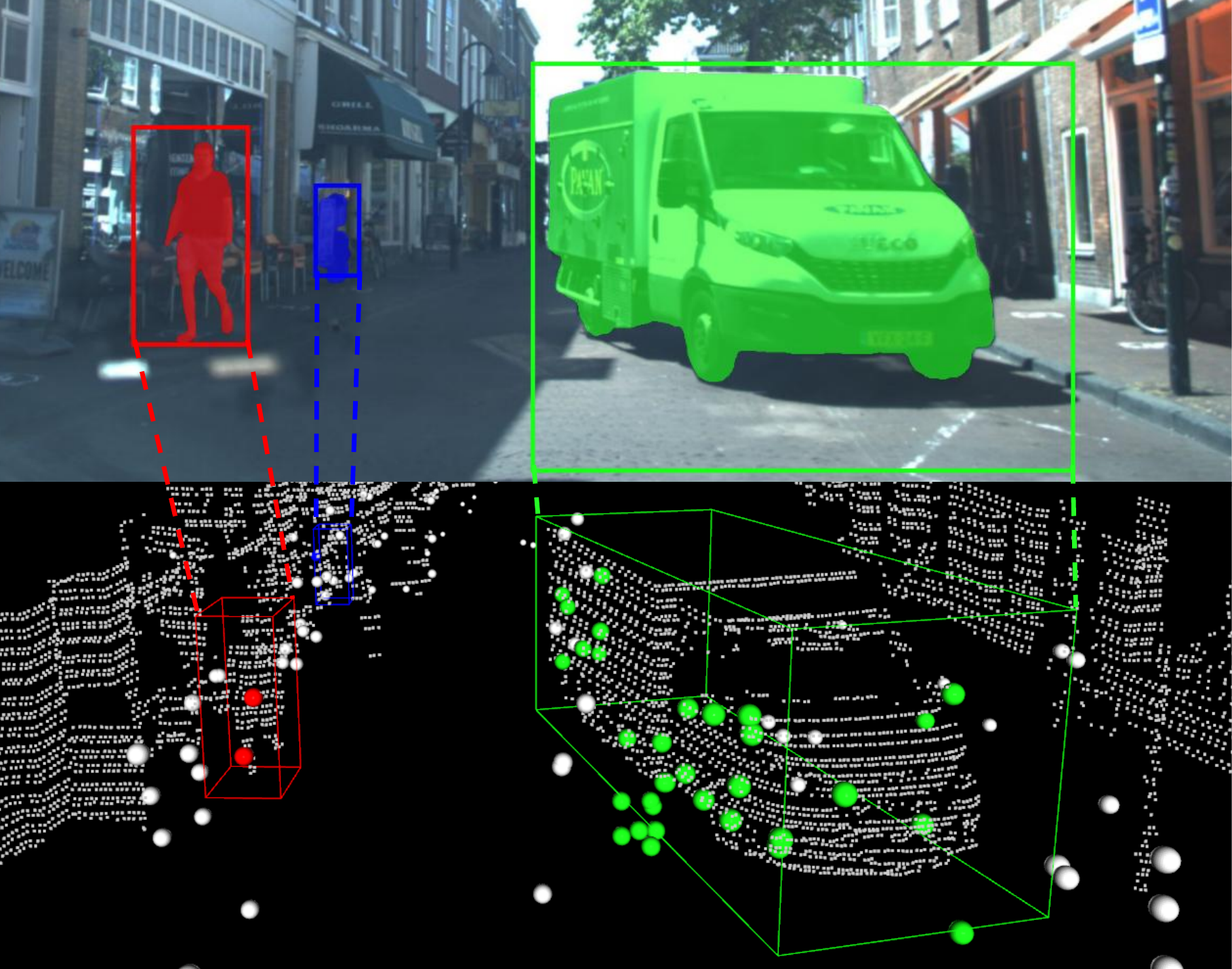}
   \caption{A schematic diagram of obtaining 3D pointwise instance labels from 2D tracking and semantic segmentation results. The smaller points are LiDAR points for auxiliary visualization, while the larger balls are radar points. Each valid radar point associated with a certain instance is painted with the same color as its corresponding 2D tracking box and instance mask. }
   \label{fig:sup_label}
\end{figure}

Given the intrinsic matrix $\Gamma_{I}$ of the camera and the extrinsic transformation matrix $\Gamma_{r \to c}$ between the radar's and camera's coordinate system, 3D radar points can be projected onto the 2D image plane:
\begin{equation}
    \label{eq:projection}
    [u, v, 1]^\text{T} = \frac{1}{z} \times \Gamma_I \times \Gamma_{r \to c} \times [x, y, z]^\text{T}~,
\end{equation}
where $x,y,z$ are the 3D spatial position of each radar point and $u,v$ is the corresponding 2D pixel coordinate. Each radar point is thereby mapped to the image plane to be associated with a tracked instance or the background and assigned with corresponding instance label or background label. Fig.~\ref{fig:sup_label} shows an example of retrieving instance labels for radar points based on the 3D-to-2D correspondence.

\begin{figure*}[t!]
  \centering
\includegraphics[width=0.99\linewidth]{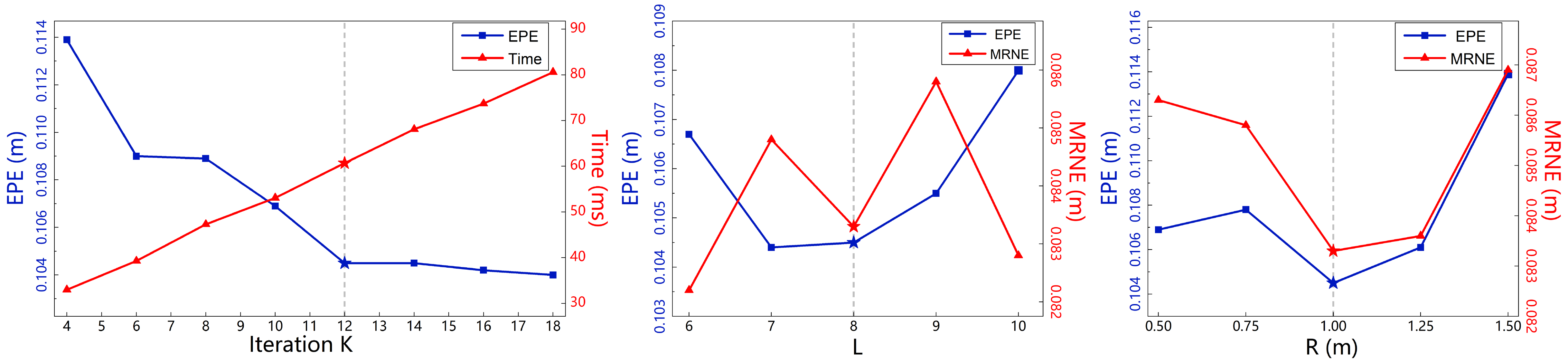}
   \caption{Ablation on iteration steps $\mathrm{K}$ and ball query hyperparameters $\mathrm{L}$ and $\mathrm{R}$. When $\mathrm{L}$ varies, $\mathrm{R}=1\mathrm{m}$; when $\mathrm{R}$ varies, $\mathrm{L}=8$.}
   \label{fig:line}
\end{figure*}

\begin{table*}[ht!]
\caption{\textbf{3-way EPE Evaluation and Runtime Comparison on VoD validation set.} IterFlow$^{\dag}$ is a lower version of our IterFlow, which uses $\mathrm{K}=4$ scene flow iterations during both training and evaluation instead of the original setting of $\mathrm{K}=12$.  The (3D Label-free) column indicates whether the network require pseudo 3D scene flow labels derived from dense 3D LiDAR point clouds with off-the-shelf 3D tracking models or ground-truth 3D tracking boxes from the dataset.}
\centering
\resizebox{0.95\textwidth}{!}{
\begin{tabular}{l|c|c|c|ccccc|c}
\toprule[2pt]
\multirow{3}{*}{Category} & \multirow{3}{*}{Method} &  \multirow{3}{*}{Sup.} & \multirow{3}{*}{\makecell{3D \\ Label \\ Free}}     & \multicolumn{5}{c|}{Radar}  & \multirow{3}{*}{\makecell{Time \\ (ms)}}
\\ \cline{5-9}  
&       &      &                         & \multicolumn{5}{c|}{Endpoint Error ($\downarrow$)}  &
\\ \cline{5-9}
& 
&       &                     
 & 3-way & 3D & FD & BS & FS &  
\\ 
\midrule[1pt]
\multirow{7}{*}{\makecell{LiDAR-\\based}} & Flow4D~\cite{kim2025flow4d}    & Fully  & \ding{55} 
& $0.2247$ & $0.2555$ & $0.2947$   & $0.2506$ & $0.1288$ & $82.0$
\\
& BiFlow~\cite{cheng2022bi}    & Fully  & \ding{55}
& $0.2015$ & $0.2140$ & $0.3056$   & $0.2067$ & $0.0922$ & $77.2$
\\
& DeFlow~\cite{zhang2024deflow}   & Fully  & \ding{55}  
& $0.1802$ & $0.2015$ & $0.2422$   & $0.1966$ & $0.1020$ & $44.8$
\\
& PointPWC~\cite{wu2020pointpwc}   & Fully  & \ding{55}  
& $0.1298$ & $0.1271$ & $0.2125$   & $0.1193$ & $0.0577$ & $53.2$
\\
& PV-RAFT~\cite{wei2021pv}   & Fully  & \ding{55} 
& $0.1238$ & $0.1140$ & $0.2126$   & $0.1044$ & $0.0545$ & $86.9$
\\

& PointPWC~\cite{wu2020pointpwc}   &Self  & \ding{51}   
& $0.3560$ & $0.4314$ & $0.4203$   & $0.4333$ & $0.2145$ & $52.1$
\\
& FlowStep3D~\cite{kittenplon2021flowstep3d}   &Self  & \ding{51} 
& $0.2296$ & $0.2607$ & $0.3116$   & $0.2562$ & $0.1210$ & $58.9$
\\
\midrule[1pt]
\multirow{4}{*}{\makecell{radar-\\based}} & RaFlow~\cite{ding2022self}   &Self  & \ding{51} 
& $0.2445$ & $0.2753$ & $0.3391$   & $0.2712$ & $0.1233$ & $\textbf{28.9}$
\\
& CMFlow~\cite{ding2023hidden}   & Cross  & \ding{55}  
& $0.1455$ & $0.1600$ & $0.2073$   & $0.1566$ & $0.0727$ & $29.1$
\\
& IterFlow$^{\dag}$ (ours)   & Cross  & \ding{51}  
& $0.1223$ & $0.1139$ & $0.2129$   & $0.1049$ & $0.0491$ & $33.0$
\\
& IterFlow (ours)    & Cross & \ding{51}  
& $\textbf{0.1156}$ & $\textbf{0.1045}$ & $\textbf{0.2058}$   & $\textbf{0.0952}$ & $\textbf{0.0458}$ & $60.7$
\\
\bottomrule[2pt]
\end{tabular}}
\label{tab:3way}
\end{table*}

\subsection{Hyperparameter Sensitivity Analysis.}
\label{sec:abla_iter}
As shown in Fig.~\ref{fig:line}, we conduct experiments on the number of scene flow iterations steps $\mathrm{K}$, and ball query hyperparameters, including the search radius $\mathrm{R}$ and the number of sampled points $\mathrm{L}$. 
As $\mathrm{K}$ increases, the performance of the network will improve until it approaches convergence, and the time consumption will also increase accordingly. Considering the trade-off between accuracy and efficiency, we set $\mathrm{K}=12$. As for $\mathrm{R}$ and $\mathrm{L}$, the accuracy of scene flow estimation varies with different hyperparameter configurations. We select the parameter combination that best optimizes network performance: $\mathrm{R}=1\mathrm{m}$ and $\mathrm{L}=8$, as described in Sec.~\ref{sec:imp}.

\section{Additional Experiments and Analysis}
\label{sec:exp}

\subsection{Three-way Endpoint Error Evaluation and Runtime Comparison} 
\label{sec:3way}

To comprehensively evaluate the performance of state-of-the-art scene flow estimation methods and our IterFlow, we adopt the three-way Endpoint Error (3-way EPE) for additional evaluation, which is often used in LiDAR-based cases~\cite{jund2021scalable,zhang2024deflow,zhang2024seflow,khoche2025ssf,kim2025flow4d}. The 3-way EPE computes the unweighted average EPE of points located in the Foreground Dynamic (`$\mathcal{FD}$'), Background Static (`$\mathcal{BS}$') and Foreground Static (`$\mathcal{FS}$') regions. The commonly used 3D end-point-error (3D EPE) results are also presented for comparison.

Note that existing cross-modal supervised methods require additional supervision from dense 3D LiDAR point clouds, 2D optical flow, and ego vehicle's odometry; and fully-supervised methods use scene flow ground truth for training. The remaining methods (including ours) are free of such direct 3D motion supervision.
As shown in Table~\ref{tab:3way}, although our IterFlow is only trained with three 3D label-free losses, it still surpasses cross-modal supervised CMFlow~\cite{ding2023hidden} on 3-way EPE metrics by a large margin, and even outperforms a series of fully-supervised models. This expanded experimental analysis further proves the effectiveness of our network and loss designs.

\begin{table*}[ht!]
\caption{Per-category Performance Disparities for FD Objects on VoD validation set.}
\centering
\resizebox{0.85\textwidth}{!}{
{
\begin{tabular}{c|c|c|c|c|c|c|c}
\toprule[2pt]
\multicolumn{3}{c|}{}  &\multicolumn{5}{c}{Speed Normalized EPE ($\downarrow$)} 
\\ 
\midrule[1pt]
Cat. & Method & Sup. & Mean & Car & O. V. & Pd. & W. V
\\ \hline
\multirow{7}{*}{\makecell{LiDAR-\\based}} & Flow4D~\cite{kim2025flow4d} & Fully & $1.3979$ & $1.1186$  & $1.6613$  & $1.4579$  & $1.3538$
\\
& BiFlow~\cite{cheng2022bi} & Fully & $1.1973$ & $1.1842$  & $1.1279$  & $1.2416$  & $1.2356$
\\
& DeFlow~\cite{zhang2024deflow} & Fully & $1.1508$ & $0.9907$  & $1.3374$  & $1.2063$  & $1.0690$
\\
& PointPWC~\cite{wu2020pointpwc} & Fully & $0.9919$ & $0.9154$  & $1.2938$  & $0.9768$  & $0.7817$
\\
& PV-RAFT~\cite{wei2021pv} & Fully  & $0.9306$ & $0.8949$  & $1.0119$  & $1.0117$  & $0.8041$
\\
& PointPWC~\cite{wu2020pointpwc} & Self & $1.8967$ & $1.3779$  & $2.2572$  & $2.0292$  & $1.9226$
\\
& FlowStep3D~\cite{kittenplon2021flowstep3d} & Self & $1.2397$ & $1.1386$  & $1.0888$  & $1.3335$  & $1.3981$
\\
\hline
\multirow{3}{*}{\makecell{Radar-\\based}} & RaFlow~\cite{ding2022self} & Self & $1.4237$ & $1.3608$  & $1.4364$  & $1.4597$  & $1.4378$
\\
& CMFlow~\cite{ding2023hidden} & Cross & $0.9676$ & $0.9029$  & $1.0947$  & $1.0396$  & $0.8332$
\\
& IterFlow (ours) & Cross & $0.8848$ & $0.8193$  & $0.8275$  & $1.0044$  & $0.8877$
\\
\bottomrule[2pt]
\end{tabular}}}
\label{tab:cat}
\end{table*}

\begin{table*}[ht!]
\caption{\textbf{
Quantitative Evaluation on MAN TruckScenes val set.
} 
In the Category (Cat.) column, existing methods are classified depending on the input modality used in their original work. In the Supervision (Sup.) column, the methods are categorized as fully supervised (Full), self-supervised (Self), or cross-supervised (Cross). * denotes using the ground truth 3D track IDs instead of pointwise instance IDs derived from 2D images for instance-aware loss calculation.
}
\centering
    \resizebox{0.99\textwidth}{!}{
    \large
\begin{tabular}{lc|c|cccccc|c|c}
\toprule[2pt]
& \multicolumn{2}{l|}{} 
& \multicolumn{6}{c|}{Overall}             & \multicolumn{1}{c|}{Moving}                                          & \multicolumn{1}{c}{Static}                   \\ 
\hline
\multicolumn{1}{c|}{Cat.}              & \multicolumn{1}{c|}{Method}  & Sup.                  &  & \quad EPE [m]$\downarrow$ \quad                              & \quad AccS [\%]$\uparrow$ \quad                            & \quad AccR [\%]$\uparrow$ \quad                            & \quad \enspace RNE [m]$\downarrow$ \enspace \quad &    & \quad MRNE [m]$\downarrow$ \quad                               &  \quad SRNE [m]$\downarrow$   \quad                      \\ 
\midrule[1pt]
\multicolumn{1}{l|}{\multirow{3}{*}{\makecell{LiDAR-\\based}}} & \multicolumn{1}{l|}{Flow4D~\cite{kim2025flow4d}} & Full  &  
& $0.2368$    & $19.07$     & $51.93$    & $0.0934$    &  &  
 $0.1077$      &   $0.0866$       
\\
\multicolumn{1}{l|}{} & \multicolumn{1}{l|}{DeFlow~\cite{zhang2024deflow}}  & Full  &  
& $0.1984$    & $38.13$     & $63.48$    & $0.0783$    &  &  
 $0.0748$      &   $0.0745$       
\\
\multicolumn{1}{l|}{} & \multicolumn{1}{l|}{PointPWC~\cite{wu2020pointpwc}}   & Full  & 
& $0.1151$    & $\textbf{70.63}$    & $\textbf{86.23}$    & $0.0454$    &  &  
 $\textbf{0.0639}$      &   $0.0413$     
\\
\midrule[1pt]
\multicolumn{1}{l|}{\multirow{4}{*}{\makecell{Radar-\\based}}}   &\multicolumn{1}{l|}{RaFlow~\cite{ding2022self}}    & Self  &  
& $0.4537$     & $2.59$      & $13.43$    & $0.1789$    &  &   $0.1334$      &   $0.1710$   
\\
\multicolumn{1}{l|}{} & \multicolumn{1}{l|}{CMFlow~\cite{ding2023hidden}}   & Full\&Cross  &  
& $0.1342$     & $38.20$      & $73.17$    & $0.0531$   &  &  
 $0.0253$      &   $0.0522$       
\\
\multicolumn{1}{l|}{} & \multicolumn{1}{l|}{IterFlow (ours)}   &Cross  &  
& $0.1103$     & $64.30$      & $83.00$    & $0.0435$   &  &  
 $0.0680$      &   $0.0389$       
\\
\multicolumn{1}{l|}{} & \multicolumn{1}{l|}{IterFlow* (ours)}  & Cross  &  
& $\textbf{0.0996}$    & $68.53$      & $84.65$     & $\textbf{0.0393}$    &  &  
 $0.0673$      &   $\textbf{0.0342}$    
\\
\bottomrule[2pt]
\end{tabular}
}
\label{tab:man_truck}
\end{table*}

Evaluated on a single NVIDIA A40 GPU, the time required for all methods to process a pair of input radar frames is recorded in Table~\ref{tab:3way}.
As described in Sec.~\ref{sec:imp} and Sec.~\ref{sec:abla_iter}, our best IterFlow sets the number of scene flow iterations steps to $\mathrm{K}=12$ for both network training and inference, which still maintains good time efficiency, as shown in the runtime column of Table~\ref{tab:3way}. We also provide the evaluation results of a lower version of IterFlow, which sets the number of iteration steps as $\mathrm{K}=4$, denoted as IterFlow$^{\dag}$ in Table~\ref{tab:3way}. It can be seen that the runtime required for our IterFlow$^{\dag}$ is similar as CMFlow~\cite{ding2023hidden}, and both methods have achieved real-time performance; however, our IterFlow$^{\dag}$ outperforms CMFlow on the 3-way EPE and 3D EPE metrics, with performance improvements of 15.9\% and 28.8\%, respectively. The experimental results in Table~\ref{tab:3way} show that IterFlow$^{\dag}$ brings a significant performance improvement in the static areas, but the EPE in the foreground dynamic (FD) region increases slightly due to the lack of direct 3D motion supervision and the limited iteration steps.

\subsection{Foreground Category Performance Analysis}
We adpot the Bucketed Normalized EPE metric~\cite{khatri2024can} to ensure that objects with different speeds are fairly evaluated. The dynamic normalized EPE is a ratio as the EPE has been normalized by speed. As shown in Table~\ref{tab:cat}, fine-grained analysis is conducted on individual classes, including Car, Other Vehicles (O. V.), Pedestrian (Pd), and Wheeled VRU (W. V.) on VoD validation dataset. The experimental results indicate that in traffic scenarios, small targets with low-speed (\textit{e.g.} pedestrians) are more challenging for scene flow estimation than other kinds of objects.

\subsection{Supplementary Experimental Results on MAN TruckScenes Dataset}
To broaden the evaluation scope, we conduct supplementary experiments on the latest MAN TruckScenes dataset, which contains 747 scenes under diverse environmental conditions, including fog, rain, snow, daytime, and nighttime. Note that MAN TruckScenes provides radar data sampled at 10 Hz but manual annotations only at 2 Hz, so intermediate sweeps are unlabeled and cannot be directly used for standard scene flow evaluation. Since ground truth scene flow is not provided in the original dataset, we generate scene flow labels based on the annotated tracking boxes for the keyframes in its official training and validation splits.

As shown in Table~\ref{tab:man_truck}, all scene flow estimation methods are trained from scratch on the MAN TruckScenes training set and evaluate on the validation set.  To train CMFlow~\cite{ding2023hidden} on the MAN TruckScenes dataset, we generate extra required 2D optical flow labels by adopting their officially released pretrained 2D optical flow estimation model and we directly adopt ground truth 3D scene flow labels to provide full supervision for CMFlow's training, since no 3D tracking results from pretrained model on the MAN TruckScenes dataset are provided. The training settings of other methods on the MAN TruckScenes dataset remain the same as their original settings on the VoD dataset. Experimental results show that our method still achieves the best performance on overall EPE metric and obtains performance comparable to fully supervised methods on the challenging MAN TruckScenes dataset, which demonstrates the effectiveness of our design. Moreover, compared to the last two rows of Table~\ref{tab:loss}, the performance gap between IterFlow* and IterFlow on the MAN TruckScenes dataset is more pronounced than that on the VoD dataset. This is because unlike the VoD dataset whose weather conditions are generally good, MAN TruckScenes covers a wide range of environmental conditions, making it more difficult for 2D models to generate accurate instance labels during training.

\begin{table}[h!]
\caption{Detection Performance of different versions of YOLO11 Model on the COCO validation set.}
\resizebox{0.47\textwidth}{!}{
\centering
{
\begin{tabular}{c|c|c|c|c}
\toprule[2pt]

Model & \makecell{mAP \\50-95} & \makecell{Runtime\\(ms)} & \makecell{params\\ (M)} & \makecell{ FLOPs\\ (B)}
\\ 
\midrule[1pt]
YOLO11m & $51.5$  & $4.7$  & $20.1$  & $68.0$
\\
YOLO11l & $53.4$  & $6.2$  & $25.3$  & $86.9$
\\
YOLO11x & $54.7$  & $11.3$  & $56.9$  & $194.9$
\\
\bottomrule[2pt]
\end{tabular}}}
\label{tab:yolo}
\end{table}

\begin{table}[ht!]
\caption{2D Zero-shot Segmentation Performance of SAM with different scales of Vision Transformer(ViT) backbone on the COCO dataset.}
\resizebox{0.49\textwidth}{!}{
\centering
{
\begin{tabular}{c|c|c|c|c|c}
\toprule[2pt]

\makecell{Backbone \\ Version} &\makecell{mIoU \\(\%)}  & \makecell{mAP\\@0.5} & \makecell{mAP \\ @0.75}  & \makecell{FPS} & \makecell{Params\\ (B)}
\\ 
\midrule[1pt]
ViT-B (Base) & $74.3$  & $78.2$  & $71.6$  & $22.2$ &  $375$M
\\
ViT-L (Large) & $76.8$  & $80.9$  & $74.5$  & $12.8$ & $1.25$G
\\
ViT-H (Huge) & $78.2$  & $82.5$  & $76.8$  & $8.0$&   $2.56$G
\\
\bottomrule[2pt]
\end{tabular}}}
\label{tab:SAM}
\end{table}

\begin{table*}[h!]
\centering
\caption{Comparison of IterFlow's Performance with 2D semantic guidance from different versions of off-the-shelf 2D tracking and segmentation models on VoD validation set.}
    \resizebox{0.99\textwidth}{!}{
    \large
\begin{tabular}{cc|cccccc|ccc|ccc}
\toprule[2pt]
& \multicolumn{1}{c|}{} & \multicolumn{6}{c|}{Overall}             & \multicolumn{3}{c|}{Moving}                                          & \multicolumn{3}{c}{Static}                                          \\ \hline
\multicolumn{1}{c|}{\makecell{YOLO11 \\Model}}               & \makecell{SAM \\Model}                  &  & \quad EPE [m]$\downarrow$ \quad                              & \quad AccS [\%]$\uparrow$ \quad                            & \quad AccR [\%]$\uparrow$ \quad                            & \quad \enspace RNE [m]$\downarrow$ \enspace \quad &  &  & \quad MRNE [m]$\downarrow$ \quad                             &  &  & \quad SRNE [m]$\downarrow$ \quad        &                    \\ 
\midrule[1pt]
\multicolumn{1}{c|}{YOLO11m} & H  &  
& $0.1025$    & $33.90$      & $64.91$     & $0.0413$    &  &  
& $0.0850$    &  &  & $0.0369$     &  
\\
\multicolumn{1}{c|}{YOLO11l} & H  &  
& $0.1045$    & $33.40$      & $63.75$     & $0.0420$    &  &  
& $0.0833$    &  &  & $0.0380$     &  
\\
\multicolumn{1}{c|}{YOLO11x} & H  &  
& $0.1031$    & $34.42$      & $65.46$     & $0.0415$    &  &  
& $0.0877$    &  &  & $0.0371$     &  
\\

\midrule[1pt]
\multicolumn{1}{c|}{YOLO11l} & B  &  
& $0.1039$    & $33.78$      & $64.46$     & $0.0418$    &  &  
& $0.0880$    &  &  & $0.0374$     &  
\\
\multicolumn{1}{c|}{YOLO11l} & L  &  
& $0.1040$    & $33.22$      & $64.29$     & $0.0419$    &  &  
& $0.0883$    &  &  & $0.0373$     &  
\\
\multicolumn{1}{c|}{YOLO11l} & H  &  
& $0.1045$    & $33.40$      & $63.75$     & $0.0420$    &  &  
& $0.0833$    &  &  & $0.0380$     &  
\\
\bottomrule[2pt]
\end{tabular}
}
\label{tab:2d_ins}
\end{table*}

\begin{figure*}[h!]
  \centering
   \includegraphics[width=0.95\linewidth]{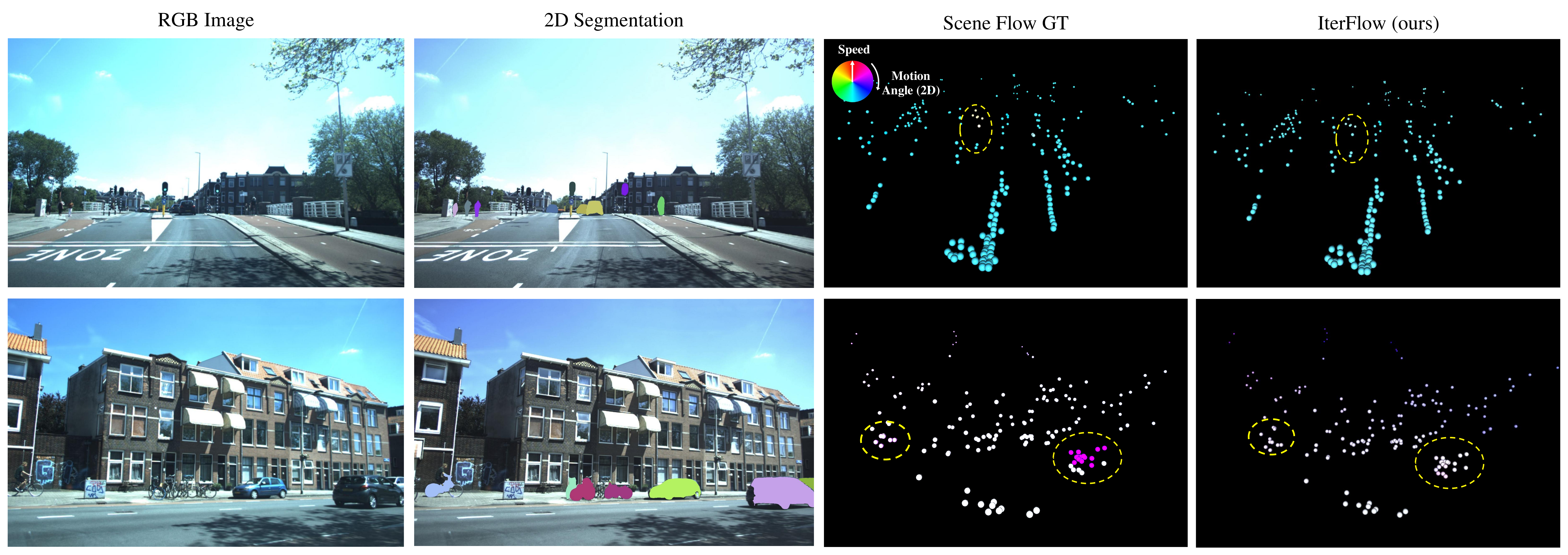}
   \caption{Visualization of failure cases on VoD validation set. Each row displays a driving scenario and regions with large scene flow estimation errors are highlighted with yellow circles.}
   \label{fig:failure}
\end{figure*}

\subsection{Supplementary Experimental Results on Various Versions of 2D Tracking and Segmentation Models}
\label{sec:sup_2d}
As described in Sec.~\ref{sec:imp}, we use the officially released YOLO11-l~\cite{khanam2024yolov11} model and SAM~\cite{kirillov2023segment} with huge version of Vision Transformer (ViT-H) backbone for 2D instance label generation in the main experimental section of this paper. As shown in Table~\ref{tab:yolo} and Table~\ref{tab:SAM}, the official repository of YOLO11 and SAM also provide various versions of pretrained models with different parameter size. The YOLO11 models are trained on the COCO dataset~\cite{lin2014microsoft} and the SAM models are trained on their SA-1B dataset, and more details are given in their official documents.

As shown in Table~\ref{tab:2d_ins}, we provide supplementary scene flow estimation results of IterFlow with 2D instance-level guidance from different combinations of off-the-shelf 2D models. It can also be observed that current commonly used 2D models can effectively provide weak supervision signals for our 4D radar scene flow estimation method and the network performance is generally robust as the combination of 2D models varies.

\section{Failure Cases and Limitation Analysis}
\label{sec:sup_vis}

As shown in Fig.~\ref{fig:failure}, we provide qualitative visualization results of failure cases for our method on VoD validation set. It can be observed that although reliable 2D instance segmentation masks have been provided by off-the-shelf 2D models, the network may fail in difficult situations, which indicates that our method is mainly limited by the sparse and noisy nature of radar data itself, rather than insufficient support from 2D semantics.

In the scene of the first row, the network incorrectly estimates the scene flow of a small number of sparse points on distant pedestrians. This is because the radar points are extremely sparse in distant areas and very few valid points are associated with difficult dynamic foreground objects; In this case, very small noise may prevent these foreground radar points from being projected onto the correct pixels in the 2D image, further impacting the optimization of predicted flows in such challenging situations.

In another scene below, the network fails to estimate the scene flows on the left cyclist and the right vehicle. Handling the large displacements of dynamic foreground objects between frames is naturally hard for scene flow estimation tasks, since the cross-frame feature correlation for flow embedding usually relies on nearest neighbor search between the point clouds. Such challenges become more prominent when it comes to radar scene flow estimation in real-world traffic scenarios, because the lack of reliable correspondences for matching instance pairs across frames and radar data is more sensitive to abnormal deformations and random noise due to the sparsity.


\end{document}